\documentclass[journal]{IEEEtai}

\usepackage[colorlinks,urlcolor=blue,linkcolor=blue,citecolor=blue]{hyperref}
\usepackage{color,array}
\usepackage{graphicx}
\usepackage{tabularx}
\usepackage{booktabs}
\usepackage{bm}
\let\labelindent\relax
\usepackage{enumitem}
\usepackage{amsmath}
\usepackage{amssymb}
\usepackage{xspace}
\usepackage{accents}
\usepackage{algorithm}
\usepackage{algpseudocode}

\newcommand{\Design}{$\mathsf{HEAL}$\xspace}
\newcommand{\Designn}{$\mathsf{HEALneural}$\xspace}
\newcommand{\Designd}{$\mathsf{HEALdiverse}$\xspace}
\newcommand{\bcs}[1]{\bm{\mathcal{#1}}}
\newcommand{\brackets}[1]{\left({#1}\right)}

\DeclareMathOperator*{\argmax}{argmax}

\begin{document}

\title{HEAL: Brain-inspired \underline{H}yperdimensional \underline{E}fficient \underline{A}ctive \underline{L}earning}

\author{Yang Ni, Zhuowen Zou, Wenjun Huang, Hanning Chen, William Youngwoo Chung, Samuel Cho, \\ Ranganath Krishnan, Pietro Mercati, \IEEEmembership{Member, IEEE}, and Mohsen Imani, \IEEEmembership{Member, IEEE}
\thanks{This work was supported in part by DARPA Young Faculty Award, National Science Foundation \#2127780, \#2319198, \#2321840, \#2312517, and \#2235472, Semiconductor Research Corporation (SRC), Office of Naval Research through the Young Investigator Program Award, and grants \#N00014-21-1-2225 and \#N00014-22-1-2067, the Air Force Office of Scientific Research, grants \#FA9550-22-1-0253, and generous gifts from Cisco.}
\thanks{Yang Ni, Zhuowen Zou, Wenjun Huang, Hanning Chen, William Youngwoo Chung, Samuel Cho, and Mohsen Imani are with the Department of Computer Science, University of California Irvine, Irvine, CA 92697 USA (e-mail: \{yni3, zhuowez1, wenjunh3, hanningc, chungwy1, samuelc7, m.imani\}@uci.edu).}
\thanks{Ranganath Krishnan and Pietro Mercati are with the Intel Labs, Hillsboro, OR 97124 USA (e-mail:\{ranganath.krishnan, pietro.mercati\}@intel.com).}}


\maketitle

\begin{abstract}
Drawing inspiration from the outstanding learning capability of our human brains, Hyperdimensional Computing (HDC) emerges as a novel computing paradigm, and it leverages high-dimensional vector presentation and operations for brain-like lightweight Machine Learning (ML). 
Practical deployments of HDC have significantly enhanced the learning efficiency compared to current deep ML methods on a broad spectrum of applications. 
However, boosting the data efficiency of HDC classifiers in supervised learning remains an open question.

In this paper, we introduce Hyperdimensional Efficient Active Learning (\Design), a novel Active Learning (AL) framework tailored for HDC classification. \Design proactively annotates unlabeled data points via uncertainty and diversity-guided acquisition, leading to a more efficient dataset annotation and lowering labor costs.
Unlike conventional AL methods that only support classifiers built upon deep neural networks (DNN), \Design operates without the need for gradient or probabilistic computations. This allows it to be effortlessly integrated with any existing HDC classifier architecture.
The key design of \Design is a novel approach for uncertainty estimation in HDC classifiers through a lightweight HDC ensemble with prior hypervectors. Additionally, by exploiting hypervectors as prototypes (i.e., compact representations), we develop an extra metric for \Design to select diverse samples within each batch for annotation.
Our evaluation shows that \Design surpasses a diverse set of baselines in AL quality and achieves notably faster acquisition than many BNN-powered or diversity-guided AL methods, recording 11$\times$ to 40,000$\times$ speedup in acquisition runtime per batch.
\end{abstract}



\section{Introduction}

\IEEEPARstart{T}{he} unparalleled performance of modern Machine Learning (ML) techniques such as Deep Neural Networks (DNN) is underpinned by the availability of vast and diverse data sources, facilitating ML applications in myriad real-world scenarios. Although beneficial for complex tasks, DNNs suffer from computational inefficiencies primarily due to their large model sizes and resource-demanding learning process~\cite{brown2020language,touvron2023llama}. Consequently, DNNs are less suitable for edge and real-time applications. Conversely, an emerging and brain-inspired computing paradigm named HyperDimensional Computing (HDC) has shed light on a more lightweight path toward learning and reasoning~\cite{kanerva2009hyperdimensional,hernandez2021onlinehd,poduval2022graphd,barkam2023hdgim}. 

The human brain can process and retrieve information with remarkable robustness and efficiency based on neural signals with high dimensions~\cite{babadi2014sparseness}. In HDC, this motivates researchers to encode inputs to high-dimensional vector representations, i.e., hypervectors~\cite{plate1995holographic}. Building upon this hypervector encoding, previous works provide a well-defined set of hypervector operations for symbol representation, concept manipulation, and crucially, model learning~\cite{frady2022computing,kanerva2009hyperdimensional}. These operations are also designed to closely mimic human brain functionalities~\cite{camina2017neuroanatomical}, fostering efficient learning, memorization, and information retrieval. In real-world implementations, HDC distinguishes itself from DNN by leveraging its hardware-friendly and straightforward mathematical operations to ensure lightweight and parallelizable processing~\cite{chen2023hypergraf,chen2022darl,ni2022algorithm}, making it particularly suitable for swift online learning with limited computing resources~\cite{hernandez2021onlinehd, ni2022hdpg}.

In the prevalent supervised learning scenario, a faster learner like HDC is undoubtedly important for lowering the learning cost. Nonetheless, the cost of acquiring labeled data cannot be ignored given the high costs and labor intensity of annotation. This poses a major challenge nowadays since the success of current ML algorithms is coupled with the cost of obtaining sufficient high-quality labeled data beforehand. Especially, with the model complexity growing, learning supervised DNN becomes significantly more data-demanding. While HDC surpasses DNN in learning efficiency, especially regarding model training iterations and runtime, it still significantly benefits from larger labeled datasets to achieve superior learning quality~\cite{ni2022algorithm,hernandez2021onlinehd}. Therefore, it is worthwhile investigating lightweight HDC learning with improved data efficiency.

We will focus on a specific strategy for improving data efficiency in this paper, namely Active Learning (AL). AL is a technique that reduces the annotation cost in data-centric and expert-driven supervised ML tasks. It proactively identifies unlabeled data that is deemed beneficial for subsequent training~\cite{houlsby2011bayesian}. This contrasts with conventional supervised ML where the model passively learns from existing labeled data; yet, not all data points are equally important and indiscriminately labeling data points could be a waste of precious annotation budget (both time and financial). Therefore, prior research has introduced various AL strategies for data-intensive DNNs to prioritize annotating data points that are more informative than others. Model uncertainty quantization~\cite{ash2019deep,hsu2015active}, representative data sampling~\cite{sener2017active}, and information-theory-based AL~\cite{kirsch2019batchbald} rank among the most widely applied methods. 

Most AL methods are designed to work side by side with DNNs, utilizing the gradient information, embeddings, or output logits of neural networks to shape the AL acquisition strategy~\cite{ash2019deep,hsu2015active,wang2014new,kirsch2019batchbald}. However, having to rely on inefficient DNN backbones means that their overall efficiency is limited. Apart from this, existing AL methods also show significant overhead, in terms of how long it takes to acquire a batch of new data points. For example, representative sampling methods require pair-wise comparison throughout the dataset and sequential greedy acquisition~\cite{sener2017active}; both incur high computation overhead and poor scalability. Uncertainty-based and information-theoretical AL methods leverage Bayesian Neural Networks (BNN) to properly quantify model uncertainty and predictive probability. However, it is well known that BNNs have even worse learning and inference efficiency when compared to regular DNNs~\cite{graves2011practical,hoffman2013stochastic,blundell2015weight,gal2016dropout}. Existing BNN methods mainly rely on variational inference to approximate the model posterior and the marginal predictive distribution~\cite{blundell2015weight,graves2011practical}. These methods vary in their implementation of variational inference and generate final predictive probability, and thus they also have different sources of inefficiency.

In this paper, we present \Design, an AL methodology that leverages brain-inspired hypervector operations to further refine HDC-based ML, enhancing sample efficiency and addressing the limitations of existing AL methods. Our contributions are summarized as follows:
\begin{itemize}[leftmargin=*]
    \item To the best of our knowledge, \Design is the first AL algorithm specifically designed for HDC-based ML. Prior HDC algorithms require the training dataset to be as complete as possible to ensure the highest learning quality. However, \Design proactively annotates unlabeled data points via uncertainty and diversity-guided acquisition, leading to a more efficient dataset annotation and lowering the labor cost. 
    \item Implementing AL within the HDC context presents significant challenges, as conventional approaches rely on BNNs and gradient-based learning for uncertainty estimation and diversity metrics. In contrast, \Design is gradient-free and seamlessly integrates with any pre-existing HDC classifier architecture. 
    \item Within the framework of \Design, we introduce a novel approach for uncertainty estimation in HDC classifiers through a lightweight HDC ensemble with prior hypervectors. The AL acquisition metric is based on average similarity margins across sub-models. Furthermore, leveraging hypervector memorization, we develop an extra metric for \Design to acquire diverse samples in batch-mode AL.
    \item Comprehensive comparison reveals that \Design outperforms in terms of AL quality and data efficiency against a diverse set of baselines on four distinct datasets. Meanwhile, \Design achieves notably faster acquisition than many BNN-powered or diversity-guided AL methods, recording 11$\times$ to 40,000$\times$ speedup in terms of acquisition runtime per batch.
\end{itemize}

As for the organization of this paper: In Section~\ref{sec:background}, we summarize the prior works, challenges, and necessary concepts regarding AL, Bayesian inference, and HDC. In Section~\ref{sec:UE}, we discuss our design in \Design to enable uncertainty estimation in HDC classification algorithms, compare several algorithm design choices, and introduce techniques to enhance the overall efficiency. We present the \Design AL algorithm with the proposed diversity metric in Section~\ref{sec:AL} and evaluate its performance against several baselines in Section~\ref{sec:Ex}.

\section{Background \& Related Works} \label{sec:background}

In this section, we formally define the problem setting for AL and cover the related AL notations that will be used in this paper (Section~\ref{sec:prob_setting}). Since a great many AL methods are powered by Bayesian inference, we introduce the necessary backgrounds for Bayesian inference methods in Section~\ref{sec:bayesian_background}, especially ones that are related to BNNs. Apart from AL techniques using BNNs, we also consider general AL methods used in deep learning and analyze the challenges in existing algorithm designs (Section~\ref{sec:challenges}). Last but not least, for readers that are not familiar with HDC, we include a concise but comprehensive introduction as well as representative related works in Section~\ref{sec:hdc_background}.

\subsection{Problem Settings for Active Learning} \label{sec:prob_setting}

In this paper, we focus on a multi-class classification problem, with ${\bcs{X}}$ being the instance space and $\bcs{Y}$ being the label space. Suppose that there are $C$ different classes in $\bcs{Y}$: $\{1,\dots,C\}$. We are interested in a classifier that predicts the label of each instance, which can be regarded as a mapping function $\bm{F_{\omega}}: \bcs{X}\rightarrow\bcs{Y}$. $\bm{\omega}$ stands for the classifier parameters. For the case of AL, we assume that the classifier is trained in a supervised way; however, instead of having a large labeled dataset in the beginning, the classifier starts training with a small labeled training set $\bcs{D}_{tr}$ and the size of training set will gradually growing as the AL algorithm kicked in. This particular training procedure is known as pool-based AL since there is a relatively large pool of unlabeled data $\bcs{D}_{pool}$. As regular supervised learning, a testing or validation dataset $\bcs{D}_{te}$ is used for evaluation. 

\begin{figure}[t]
    \centering
    \includegraphics[width=1\columnwidth]{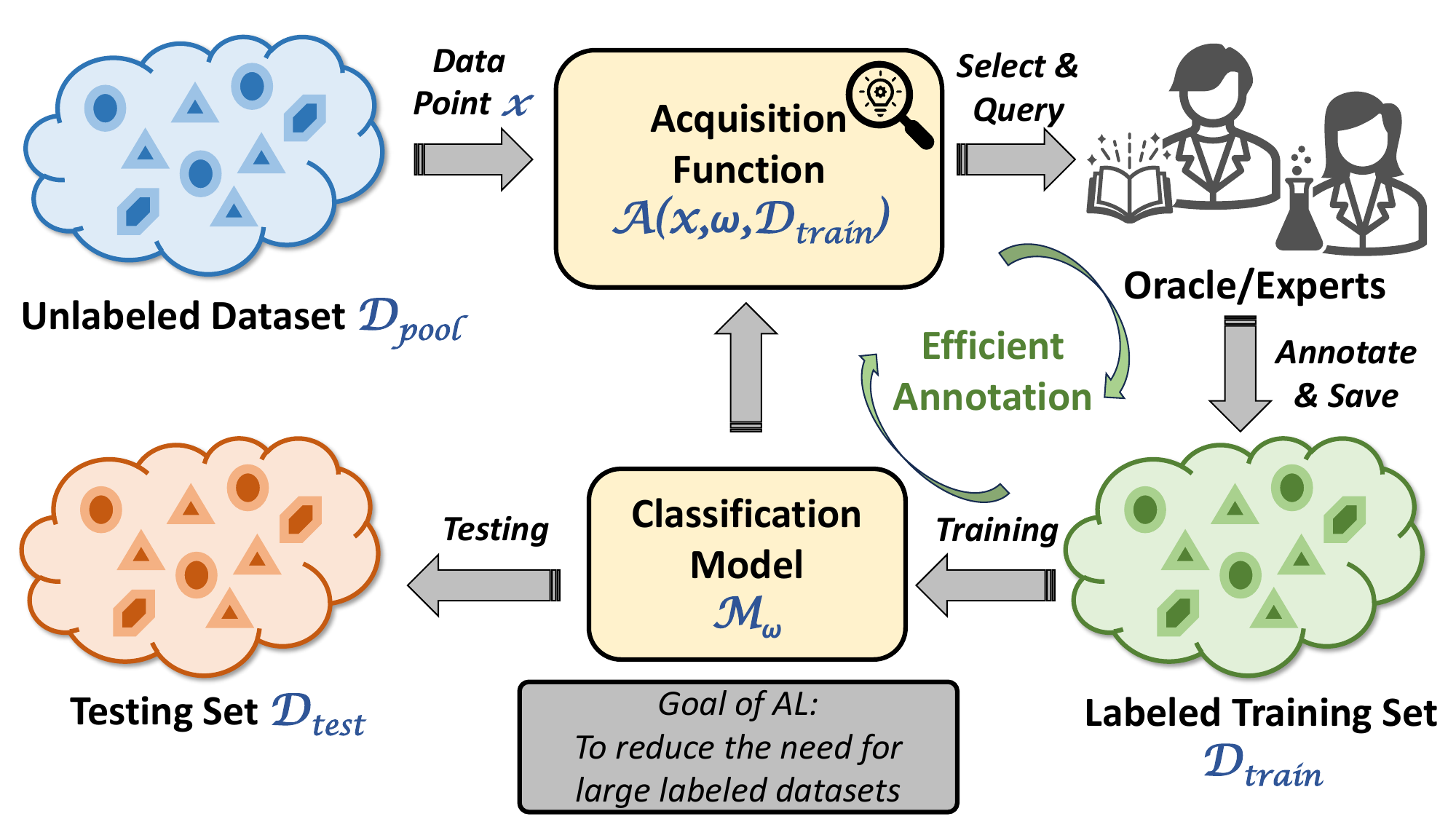}
    \caption{Outline of active learning in supervised classification.}
    \label{fig:al_overview}
\end{figure}

Fig.~\ref{fig:al_overview} presents the procedure of AL, which includes an oracle to provide extra labeled data points. The oracle can give the correct label for every data point in $\bcs{D}_{pool}$ as long as the annotation request is made from AL. The labeled data points are then removed from $\bcs{D}_{pool}$ and added to the training dataset $\bcs{D}_{tr}$. The goal of AL, in short, is to minimize the number of annotation requests (also the size of the labeled dataset) while ensuring satisfying prediction accuracy. Specifically, at each step, AL uses an acquisition function $\bcs{A}$ to rank samples in $\bcs{D}_{tr}$ and select extra data that maximizes this function, given the current classifier trained using $\bcs{D}_{tr}$:
\begin{equation}
    \bm{x}_{to\_label} = \argmax_{\bm{x}'\in \bcs{D}_{pool}} \bcs{A}\left(\bm{x}',\bm{\omega},\bcs{D}_{tr}\right)
\end{equation}
In practice, AL algorithms usually acquire a batch of new data points instead of only one $\bm{x}_{to\_label}$ at each step to reduce the frequency and cost of model retraining. This can be achieved by simply acquiring top $K$ data points according to the acquisition function $\bcs{A}$. 
Sometimes, it is also required to slightly modify the acquisition function~\cite{kirsch2019batchbald} or add extra metrics~\cite{ash2019deep} to maintain the diversity in acquired batches. (more details in Section~\ref{sec:diversity_AL}).

\subsection{Bayesian Inference for DNNs} \label{sec:bayesian_background}

Bayesian inference plays a key role in uncertainty-based AL as it provides a principled way of understanding the uncertainty inherent in predictions~\cite{kirsch2019batchbald,houlsby2011bayesian}. In Bayesian inference, the neural networks are no longer modeled in static weights $\bm\omega$ but stochastically in a distribution $p\left(\bm\omega|\bcs{D}_{tr}\right)$~\cite{neal2012bayesian}. It is known as the posterior distribution whose variance quantifies the belief in DNN weights given the seen data points in training. This also means a prior belief $p\left(\bm\omega\right)$ exists before training. Using the posterior, the predictive distribution is calculated as: $p\brackets{y|\bm{x},\bcs{D}_{tr}} = \mathbb{E}_{\bm\omega\sim p\brackets{\bm\omega|\bcs{D}_{tr}}}\left[p\brackets{y|\bm{x},\bm\omega}\right]$. We can compute the posterior using the Bayes Theorem as follows:
\begin{equation}
    p\left(\bm\omega|\bcs{D}_{tr}\right) = \frac{p\left(\bcs{D}_{tr}|\bm\omega\right)p\left(\bm\omega\right)}{\int p\brackets{\bcs{D}_{tr}|\bm\omega'}p\brackets{\bm\omega'}\,d\bm\omega'}
\end{equation}
with $p\brackets{\bcs{D}_{tr}|\bm\omega}$ being the likelihood. From the equation above, it is clear that the exact posterior will be intractable in DNNs because the weight space is generally high-dimensional and takes exponential time to evaluate the integral in the denominator~\cite{blundell2015weight}.

Therefore, approximating the posterior in a more tractable way is fundamental for achieving Bayesian inference in DNNs. Variational Inference (VI) has been a widely applied method that replaces the original $p\brackets{\bm\omega|\bcs{D}_{tr}}$ with a simpler variational distribution with a tractable $q\brackets{\bm\omega|\bm\omega_{q}}$, e.g., a family of Gaussian distributions~\cite{hoffman2013stochastic,graves2011practical}. This is usually achieved by minimizing the Kullback-Leibler (KL) divergence that measures the differences between two probability distributions. By rearranging the terms in the KL divergence, we get Evidence Lower Bound (ELBO). 

Early works that tried to apply VI for DNNs built up the foundation for what we know today as Bayesian Neural Networks (BNN)~\cite{hinton1993keeping,barber1998ensemble,jordan1999introduction}. However, they faced several challenges such as the lack of support for modern DNNs and scalability to larger datasets~\cite{mackay1995probable}. More recent BNN works~\cite{graves2011practical,hoffman2013stochastic,blundell2015weight} make it practical to apply Bayesian inference for deep learning tasks, thanks to their compatibility with most DNN structures and mini-batch gradient descent optimization. 

Apart from VI-based BNNs, researchers also develop methods based on ensemble learning for uncertainty estimation in DNNs. For example, MC-Dropout~\cite{gal2016dropout} proposes to approximate Bayesian inference by adding dropout layers. 
In its design, dropout layers are enabled also in network inference instead of just during training. 
The predictive distribution and uncertainty are derived from these multiple forward passes, assuming that the predictions follow a Gaussian distribution. 
On the other hand, Deep Ensemble~\cite{lakshminarayanan2017simple} more explicitly utilizes the ensemble of DNNs with bootstrapping and adversarial training. For classification problems, the predictive probabilities are computed as the average of ensemble softmax probabilities. Prior works~\cite{gustafsson2020evaluating,ovadia2019can} also show that Deep Ensemble generally achieves the best performance on uncertainty estimation, compared to the implicit ensemble in MC-Dropout and VI-based BNN.

\subsection{Existing DNN-AL Methods and Their Challenges} \label{sec:challenges}

As is pointed out in the introduction, one motivation of this paper is that current AL methods face various kinds of inefficiencies; these challenges are the consequences of using acquisition functions with high overhead as well as relying on computation-heavy DNNs (or BNNs) for training and inference.

\noindent\textbf{Uncertainty-based AL: }Many existing AL methods are based on a certain uncertainty metric. For example, confidence sampling~\cite{ruuvzivcka2020deep} will select samples with the smallest predictive probability $p\brackets{y|\bm{x},\bcs{D}_{tr}}$ while entropy sampling~\cite{liu2017active} selects points with the highest predictive entropy values $\mathbb{H}\brackets{y|\bm{x},\bcs{D}_{tr}}$. Both confidence and entropy provide direct estimations of model uncertainty~\cite{wang2014new}. Margin sampling~\cite{roth2006margin,lv2020deep,joshi2009multi} approaches this at a slightly different angle, where it sorts samples in the AL pool according to the probability margin $p\brackets{y|\bm{x},\bcs{D}_{tr}}-p\brackets{y'|\bm{x},\bcs{D}_{tr}}$. Here $y$ stands for the predicted label and $y'$ is the class with the second largest predictive probability. The data point with the smallest margin indicates low confidence in prediction and will be annotated by experts.

However, the aforementioned methods generally have significant acquisition costs as they require BNN during uncertainty estimation. As for the BNNs mentioned above in~\ref{sec:bayesian_background}, they all show overheads in different aspects. First, to obtain the predictive distribution, methods like MC-Dropout and Bayes-by-Backprop require more than 100 forward inferences on one single test sample. The inefficient BNN inference directly leads to a higher acquisition cost since every AL step requires scoring all data in $\bcs{D}_{pool}$. The same is true for Deep Ensemble where the inference is inefficiently carried out on multiple sub-models. Second, training BNNs is notably more costly than regular DNNs using similar architectures. For example, network parameters are doubled in BNNs that parameterize the variance of weights~\cite{blundell2015weight}, and they can have poor scalability to dataset sizes~\cite{hernandez2015probabilistic}.

In addition, AL methods with only the uncertainty metric are prone to sample duplicate data points in the batch acquisition, compromising the advantage in data efficiency. The next few categories of AL methods mitigate this problem by explicitly or implicitly considering the diversity in batch acquisition. However, they have also shown significantly higher overhead in acquisition.

\noindent\textbf{Information Theory Based AL: }AL methods~\cite{houlsby2011bayesian,kirsch2019batchbald} also utilize information theory such as the mutual information (also known as the expected information gain) to guide the acquisition. Here we show the definition of mutual information in batch acquisition (i.e., BatchBALD~\cite{kirsch2019batchbald}):
\begin{equation}
\resizebox{.43\textwidth}{!}{%
    $\mathbb{I}\brackets{y_{1:K};\bm\omega|\bcs{D}_{tr}}=\mathbb{H}\brackets{y_{1:K}|\bcs{D}_{tr}}-\mathbb{E}_{p(\bm{\omega}|\bcs{D}_{tr})}\left[\mathbb{H}\brackets{y_{1:K}|\bm\omega,\bcs{D}_{tr}}\right]$%
}
\label{eq:m_info}
\end{equation}
Notice that we omit the conditioning on $\bm{x}_{1:K}$ for each term in this equation. The first term on the right side of the equation represents the overall predictive uncertainty with the unconditioned entropy while the second term is the expected conditional entropy for each sampled model. In short, this acquisition function looks for data points on which sampled models (from the posterior $p(\bm{\omega}|\bcs{D}_{tr})$) disagree with each other.

As for the drawbacks, they still suffer from the overhead of Bayesian inference and posterior computation, since the information metric selected will only be helpful and properly calibrated with Bayesian inference. In addition, computing equation~\ref{eq:m_info} in each step of batch acquisition is time-consuming and memory-heavy as the complexity grows exponentially with the batch sizes~\cite{kirsch2019batchbald}. With a batch size larger than 10, the acquisition runtime can become inhabitable for deployment on CPUs or any resource-limited hardware.

\noindent\textbf{Representative Sampling Based AL: }Algorithms in this group aim to find a subset of data points that can behave as a surrogate for the complete training dataset, which is an intuitive method to reduce the annotation cost while ensuring learning quality. The core-set method~\cite{sener2017active,geifman2017deep} achieves this goal by leveraging the geometry of the data points and minimizing the covering radius of the selected subset, where a smaller covering radius means a better surrogate. However, the computational cost remains high for this method because it requires computing pairwise distances for every point to be added in the representative subset; this limits the scalability and deployments of this method.

\noindent\textbf{Hybrid AL Methods: } Work in~\cite{hsu2015active} applies meta-learning to help balance the part that follows the representative sampling and the part for uncertainty-based sampling. Work in~\cite{huang2010active} also aims to acquire data points that are diverse and informative. A more recent work~\cite{ash2019deep} proposes BADGE, a diversity and uncertainty-guided AL framework for DNNs, that leverages the magnitude of gradient embedding as the sign for uncertainty. It then applies the k-MEANS++ algorithm~\cite{arthur2007k} on these hallucinated gradients to ensure diversity. However, the cost of running k-MEANS++ adds significant overhead to the AL acquisition process.

\subsection{Brain-inspired HDC} \label{sec:hdc_background}

The human brain is a very efficient and robust learning machine and inspires many different machine learning algorithms. HDC, as our focus in this paper, utilizes high-dimensional representations to emulate brain functionalities for ML tasks. Representations in HDC are considered holographic since the information is evenly distributed among all dimensions of the hypervector~\cite{kanerva2009hyperdimensional}; therefore, corruption or noises on some dimensions will not lead to catastrophic information loss~\cite{poduval2022adaptive}.

As the background, we start with how HDC represents symbols and memorizes concepts in a structured way. Suppose there are three different symbols $a$, $b$, and $c$, we map them to three randomly-sampled hypervector representations $\bm{h}_a$, $\bm{h}_b$, and $\bm{h}_c$ with dimensionality $D$. More specifically, every element in those hypervectors is i.i.d. usually following a symmetric, zero-mean distribution~\cite{frady2022computing}. For example, $\bm{h}$ can be bipolar high-dimensional vectors, whose elements are either +1 or -1 with equal probability~\cite{kanerva2009hyperdimensional}.  Hypervectors can represent not only individual symbols but also a combination or association of multiple symbols, which is enabled by the following three fundamental hypervector operations.

\begin{itemize}
    \item \textbf{Hypervector Similarity ($\delta$)} represents how close two hypervectors are in the hyperspace, defined based on the normalized dot product: $\delta\brackets{\bm{h}_{a},\bm{h}_{b}}=\bm{h}_{a}\cdot\bm{h}_{b}/(||\bm{h}_{a}||*||\bm{h}_{b}||)$. When $D$ is in the range of several hundred to around ten thousand, the similarity between any two random hypervectors is nearly zero (also known as near-orthogonal) because of the property of high-dimensional space.
    \item \textbf{Bundling ($\oplus$)} in HDC means the element-wise addition of two or more hypervectors. This creates a new hypervector that represents the set of hypervectors and thus remains similar to all its constituents. Suppose we have $\bm{h}_s = \bm{h}_a \oplus \bm{h}_b$, and then we check the similarity between the bundled hypervector and each symbol hypervector: $\delta(\bm{h}_s, \bm{h}_a) \approx \delta(\bm{h}_s, \bm{h}_b) \gg 0$ while $\delta(\bm{h}_s, \bm{h}_c) \approx 0$.
    \item \textbf{Binding ($\odot$)} stands for the element-wise multiplication of two hypervectors. It generates a dissimilar vector representing their association. To associate $a$ with $b$, we bind their corresponding hypervectors: $\bm{h}_{(a,b)} = \bm{h}_a \odot \bm{h}_b$. Unlike bundling, $\bm{h}_{(a,b)}$ is dissimilar to both constituents, e.g., $\delta(\bm{h}_{(a,b)}, \bm{h}_a) \approx 0$.
\end{itemize}
The bundling operation is usually used when creating HDC-based ML models such as the classifier hypervectors and regression model hypervectors. The binding operation is used when associating features and values. We will cover more details regarding the usage of hypervector operations in Section~\ref{sec:HDC_encoding}.

In the past few years, HDC has gained significant traction as an emerging computing paradigm, especially for its deployments in machine learning and reasoning tasks. Prior works have proposed HDC-based algorithms and learning frameworks for classification~\cite{heddes2023torchhd,hernandez2021onlinehd,zou2021scalable}, clustering~\cite{gupta2022store,imani2019hdcluster}, regression~\cite{hernandez2021reghd,frady2022computing}, and reinforcement learning~\cite{ni2022hdpg,ni2023efficient,issa2022hyperdimensional} problems, showing the benefit of fast convergence in learning, high power/energy efficiency, natural data reuse and acceleration on customized devices~\cite{chen2022darl,chen2023hypergraf,zou2022biohd,chen2022full}, and robustness on error-prone emerging hardware~\cite{xu2023hypermetric,barkam2023reliable}. Particularly, HDC has been successfully applied to many supervised learning tasks. For biomedical applications, HDC-based algorithms with low power consumption are aimed at solving problems like DNA sequencing~\cite{zou2022biohd,barkam2023hdgim}, health monitoring~\cite{pale2024combining,shahhosseini2022flexible, ni2022neurally}, and subject intention recognition~\cite{rahimi2020hyperdimensional}. In work~\cite{burrello2018one}, the proposed method achieves one/few-shot learning on edge devices for the task of epileptic seizure detection. In addition, HDC has also shown significantly faster learning in classifying human faces~\cite{imani2022neural}, spam texts~\cite{thapa2021spamhd}, texts\cite{shridhar2020end}, etc. More recently, researchers also proposed to incorporate uncertainty estimation into HDC-based regression via a customized HDC encoder that randomly drops dimensions~\cite{ni2023brain}. However, this method is not suitable for our use case since dropping dimensions has little impact on classification results due to the robustness of HDC-based classification.
However, this paper focuses on incorporating uncertainty estimation into HDC classifiers and designing an AL mechanism in an efficient way for enhancing HDC-based ML with better data efficiency.

\section{Estimating Uncertainty in HDC via Efficient Ensemble} \label{sec:UE}

As we discussed in previous sections, revealing the model uncertainty is the key step in many AL techniques. Therefore, the first question we ask when designing \Design is how to find out the confidence of the HDC model on its prediction. Interestingly, prior works on BNNs show that ensemble-based design not only improves neural network learning quality but also serves as a proxy for estimating predictive uncertainty~\cite{lakshminarayanan2017simple,gal2016dropout}. Motivated by this finding, we propose an efficient HDC ensemble learning algorithm that supports \Design by providing model confidence.

In this section, we first focus on our choice of HDC encoder (in Section~\ref{sec:HDC_encoding}) since it is one of the most important components in any HDC-based algorithm. Then we will cover a naive HDC ensemble learning design in Section~\ref{sec:naive_ensemble} and qualitatively analyze its performance in uncertainty estimation. In Section~\ref{sec:HDC_prior}, we further improve the naive ensemble design by injecting prior information to HDC learning. Finally in Section~\ref{sec:efficient_design}, we discuss techniques in \Design that enhance its efficiency.

\subsection{Hypervector Encoding in HDC} \label{sec:HDC_encoding}

\begin{figure}[t]
    \centering
    \includegraphics[width=1\columnwidth]{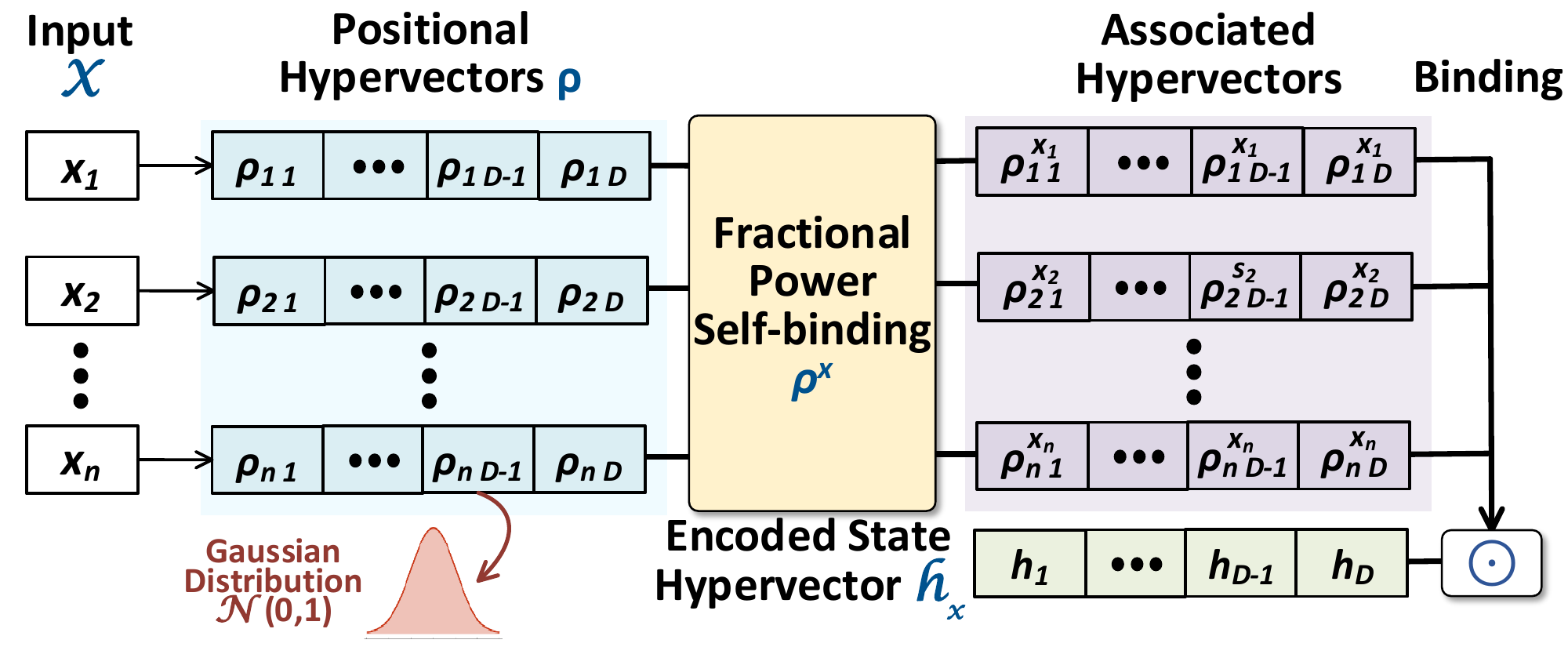}
    \caption{Encode to hypervectors via fractional power encoding.}
    \label{fig:encoding}
\end{figure}

The encoding process in HDC is essentially mapping a vector of features to the high-dimensional space or hyperspace. HDC encoders need to maintain the relationship between input features in the hyperspace, e.g., two similar inputs are mapped to hypervectors with relatively larger cosine similarity~\cite{aygun2023learning}. For \Design, we leverage the Fractional Power Encoding (FPE) to achieve a holographic reduced representation~\cite{plate1995holographic}, i.e., symbols and their structured combinations can be represented uniformly in a hyperspace. In Fig.~\ref{fig:encoding}, we provide the outline of the HDC encoding process in \Design.

We start by assuming a feature vector input of length $n$: $\bm{x}=\left[x_1,x_2,\dots,x_n\right]^{T}$, and we view each feature in $\bm{x}$ as a symbol to encode. The HDC encoder is pre-loaded with a series of positional hypervectors $\{\bm{\rho}_1,\bm{\rho}_2,\dots,\bm{\rho}_n\}$, each corresponding to one feature/position of the input feature. Elements in positional hypervectors are randomly sampled unitary phasors, i.e., $\bm{\rho}_n = \{e^{i\theta_{\bm{\rho}_n}}\}^{D}$ where $\theta_{\bm{\rho}_n}\sim\mathcal{N}(0,1)$. Thus, the randomly sampled positional hypervectors are still near-orthogonal to each other in the hyperspace. Notice that although they are no longer bipolar hypervectors, binding and bundling operations mentioned in Section~\ref{sec:hdc_background} still apply to them. 

The difference in FPE lies in the way it associates features and their values. If we use regular binding defined in Section~\ref{sec:hdc_background}, two hypervectors, one for the feature and one for its value, will be combined using element-wise multiplication (e.g., $\bm{\rho}_1\odot\bm{v}_{x_1}$). This requires value quantization such that value hypervectors $\bm{v}$ represent different discrete steps of feature values, which inevitably introduces information loss and cannot scale to data with large value ranges. FPE allows this to happen in a much finer granularity with fractional power self-binding~\cite{plate1995holographic}. Recall that binding results in a dissimilar hypervector, and thus we can encode integer feature values by repeatedly self-binding the feature positional hypervector. For example, we can encode $x_1=3$ as: $\bm{\rho}_{1}^{3}=\bm{\rho}_{1}\odot\bm{\rho}_{1}\odot\bm{\rho}_{1}$. With FPE, a feature with floating number value is encoded similarly, e.g., $x_2=5.5$ is encoded by elementwise exponential $\bm{\rho}_{2}^{5.5}=e^{i\bm{\theta}_{\bm{\rho}_{2}}*5.5}$. More generally, for the feature vector input $\bm{x}$, we have the encoding process as follows:
\begin{equation} 
\label{eq:hdc_encode}
    \bm{h}_{\bm{x}}=\phi_{encode}\brackets{\bm{x}}=\bm{\rho}_{1}^{x_1}\odot\bm{\rho}_{2}^{x_2}\cdots\odot\bm{\rho}_{n}^{x_n}=e^{i\Theta^{T}\bm{x}}
\end{equation}
where we bind all encoded features together and $\Theta$ stands for an $n\times D$ matrix with each row being the $\bm{\theta}_{\bm{\rho}_n}$ vector. Notice that $\bm{h}_{\bm{x}}\in \mathbb{C}^D$ and similarity is real-valued, the similarity computation for hypervectors with FPE is defined as:
\begin{equation}
\label{eq:sim_check}
    \delta\brackets{\bm{h}_{\bm{x}},\bm{h}_{\bm{x'}}}=\frac{real\brackets{\bm{h}_{\bm{x}}\cdot\bm{h}_{\bm{x'}}^{\dag}}}{||\bm{h}_{\bm{x}}||*||\bm{h}_{\bm{x'}}||}
\end{equation}
where $\bm{h}_{\bm{x'}}^{\dag}$ is the complex conjugate and we normalize the real part of the dot product; this is also known as the Euclidean angle for complex vectors. 

\subsection{HDC Classification and Uncertainty Estimation with Naive Ensembles} \label{sec:naive_ensemble}

BNNs with ensemble-based design have been leveraged for uncertainty-based AL since methods like MC-Dropout can estimate the model confidence via the variance in prediction~\cite{gal2017deep}. On the other hand, HDC is also compatible with ensemble learning; we can train multiple HDC sub-models, each with its own HDC encoder and bootstrapped training set, i.e., bagging. During the inference, all sub-models will contribute to the prediction in a consensus-based way.

\noindent\textbf{HDC Classification: }We will now show procedures of HDC-based classification with ensemble learning. For a dataset with $C$ classes, the HDC model is comprised of $C$ class hypervectors $\bm{M}:\{\bm{m}_1,\bm{m}_2,\cdots,\bm{m}_{C}\}$, each has the same dimensionality $D$ as $\bm{h}_{\bm{x}}$ and $\bm{\rho}$. Class hypervectors can be trained through the bundling of all encoded training samples that share the same label, i.e., they are reduced representations for classes. HDC predicts by computing similarities of the encoded query hypervector and different class hypervectors and looking for the maximum.

Assume $E$ HDC sub-models $\{\bm{M}_1,\bm{M}_2,\cdots,\bm{M}_E\}$, and the first sub-model $\bm{M}_1$ will be trained with a data point $\bm{x}$ with label $l_{true}$. We first encode this sample to hypervector $\bm{h}_{\bm{x}}$ via equation~\ref{eq:hdc_encode} and then perform a similarity check with every class hypervector in model $\bm{M}_1$ using equation~\ref{eq:sim_check}. The prediction process for $\bm{M}_1$ is shown below:
\begin{equation}
    l_{pred}=\argmax_{\bm{m}_{l}\in\bm{M}_1}\delta\brackets{\bm{h}_{\bm{x}},\bm{m}_{l}}
\end{equation}
Then, class hypervectors will be updated according to how well the model predicts. More specifically, if the prediction is not correct ($l_{pred}\neq l_{true}$), the update process for sub-model $\bm{M}_1$ is as the following:
\begin{equation}
\label{eq:correct_pred}
    \bm{m}_{l_{true}}=\bm{m}_{l_{true}}\oplus\lambda\brackets{1-\delta\brackets{\bm{h}_{\bm{x}},\bm{m}_{l_{true}}}}\bm{h}_{\bm{x}}
\end{equation}
\begin{equation}
\label{eq:wrong_pred}
    \bm{m}_{l_{pred}}=\bm{m}_{l_{pred}}\oplus\lambda\brackets{\delta\brackets{\bm{h}_{\bm{x}},\bm{m}_{l_{pred}}}-1}\bm{h}_{\bm{x}}
\end{equation}
where $\lambda$ is the learning rate. The similarity values $\delta$ in the updates function as the feedback that dynamically controls the learning rate. For instance, a $\delta\brackets{\bm{h}_{\bm{x}},\bm{m}_{l_{true}}}$ near 1 means that the class hypervector already contains the information in the query and only a slight update is needed as in equation~\ref{eq:correct_pred}; the wrong prediction may be the result of $\delta\brackets{\bm{h}_{\bm{x}},\bm{m}_{l_{pred}}}$ being incorrectly high, and therefore equation~\ref{eq:wrong_pred} will try to rectify. In addition, if the prediction is correct, then the HDC model is not updated. The process above happens separately for every sub-model during training based on bootstrapped sampling, and they are trained iteratively with batched samples.

\noindent\textbf{Ensemble Inference \& Uncertainty Estimation: }As for the inference, every sub-model gives predictions on all testing data points. And for each data point, there is an array of predicted labels: $\bm{l}_{pred}:\{l_{pred}^{\bm{M}_1},l_{pred}^{\bm{M}_2},\dots,l_{pred}^{\bm{M}_E}\}$. Regular ensemble inference uses voting to get the final prediction: $\hat{l}=\text{Mode}_{l}\brackets{\bm{l}_{pred}}$. However, here we are interested in the predictive uncertainty, which can be estimated using entropy. To compute the entropy, we first approximate the probability of every predicted class label through:
\begin{equation}
\label{eq:estimate_prob}
    p\brackets{y=l}_{l\in\bm{l}_{pred}}\approx\frac{\sum_{i=1}^{E} \left[l_{pred}^{\bm{M}_i}=l\right]}{E}
\end{equation}
where $\sum_{i=1}^{E} \left[l_{pred}^{\bm{M}_i}=l\right]$ counts the number of sub-models that predict $y=l$. Then the predictive entropy is computed as:
\begin{equation}
    \mathbb{H}\brackets{y|\bm{x},\bcs{D}_{tr}}=-\sum_{l\in\bm{l}_{pred}}p\brackets{y=l}\log p\brackets{y=l}
\end{equation}
Intuitively, a higher entropy indicates that sub-models vote for different predictions, while a lower entropy close to zero infers that sub-models agree with each other.

\begin{figure}[t]
    \centering
    \includegraphics[width=1\columnwidth]{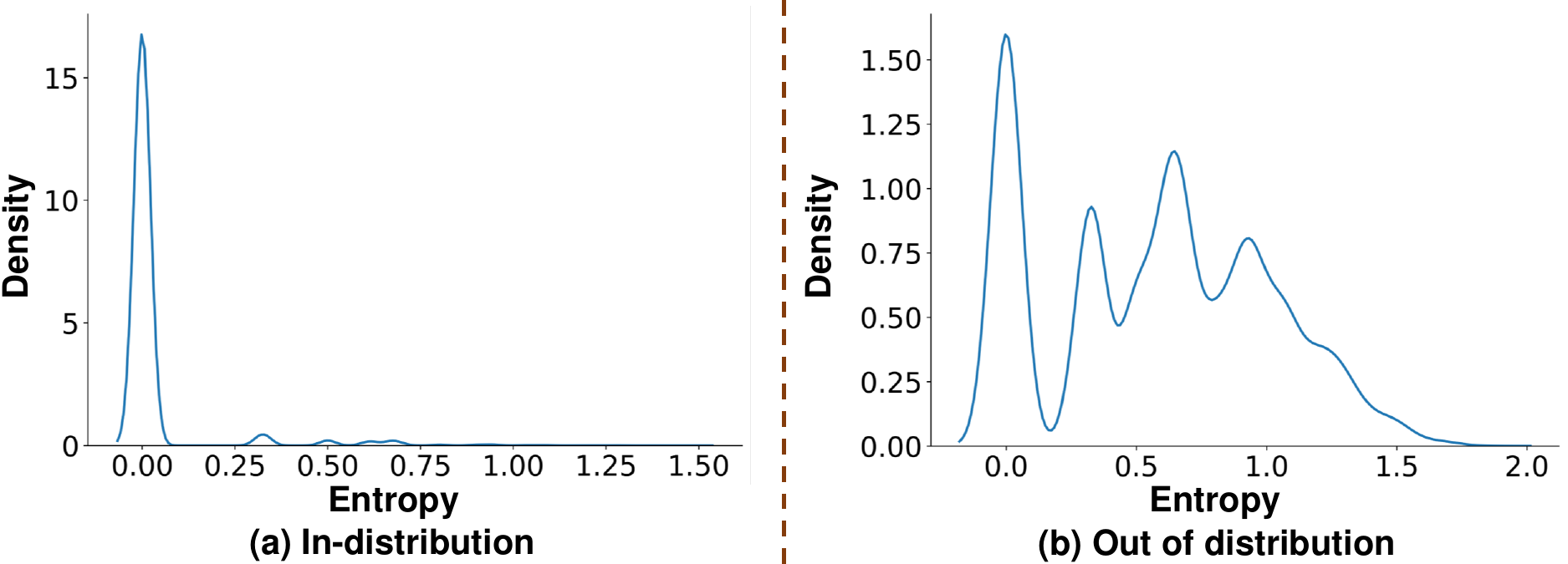}
    \caption{Histogram of predictive entropy for HDC classification with naive ensembles on (a): in-distribution testing set and (b): OOD testing set.}
    \label{fig:naive_es}
\end{figure}

To qualitatively evaluate this estimation of model uncertainty, we apply HDC ensemble training using the MNIST dataset while testing on the NotMNIST~\cite{bulatov2011notmnist} dataset, which is comprised of out-of-distribution (OOD) alphabet images with 10 classes and the same image size as MNIST. Since the ensemble model only knows handwritten digits from training, we expect the prediction probability to be closer to uniform during testing. In other words, each ensemble model will disagree with each other when evaluated on OOD samples, which results in a much larger predictive entropy.

In Fig.~\ref{fig:naive_es}, we present the histogram of predictive entropy when testing the HDC naive ensemble with samples from both MNIST and NotMNIST datasets. For the in-distribution case, the histogram in Fig.~\ref{fig:naive_es}(a) shows a high density around zero and indicates that ensemble sub-models agree with each other, i.e., lower uncertainty in prediction. As for the OOD case in Fig.~\ref{fig:naive_es}(b), it is clear by comparison that the HDC naive ensemble now shows a higher level of uncertainty as larger entropy values show up in the histogram, indicating the disagreement between sub-models for some data points. However, the quality of this uncertainty estimation is lower than expected because the highest density still occurs at zero entropy values. This means that when deploying the naive ensemble design in practice, the HDC ensemble will frequently be over-confident about its prediction, and thereby unsuitable for uncertainty-based AL.

One possible reason why the HDC naive ensemble performs poorly is that HDC sub-models fail to understand the data space from diverse perspectives, which is crucial for fostering disagreement when facing unseen data points. In fact, due to the way model hypervectors are constructed, i.e., being the superposition of seen data, even with bootstrapping, these HDC sub-models are somewhat similar after training. Therefore, there is a high chance that sub-models will predict similar labels when they actually have low confidence.

\subsection{HDC Prior Hypervectors} \label{sec:HDC_prior}

In previous HDC classification works, the model hypervectors are always initialized as zero vectors. Since these hypervectors are considered as the memory and prototype of data points of different classes, starting with zero values indicates no prior knowledge about the task or classes. Notice that this is different from DNNs with back-propagation training, as the zero initialization for neurons leads to uninformative gradients and poor learning results.

\begin{figure}[t]
    \centering
    \includegraphics[width=1\columnwidth]{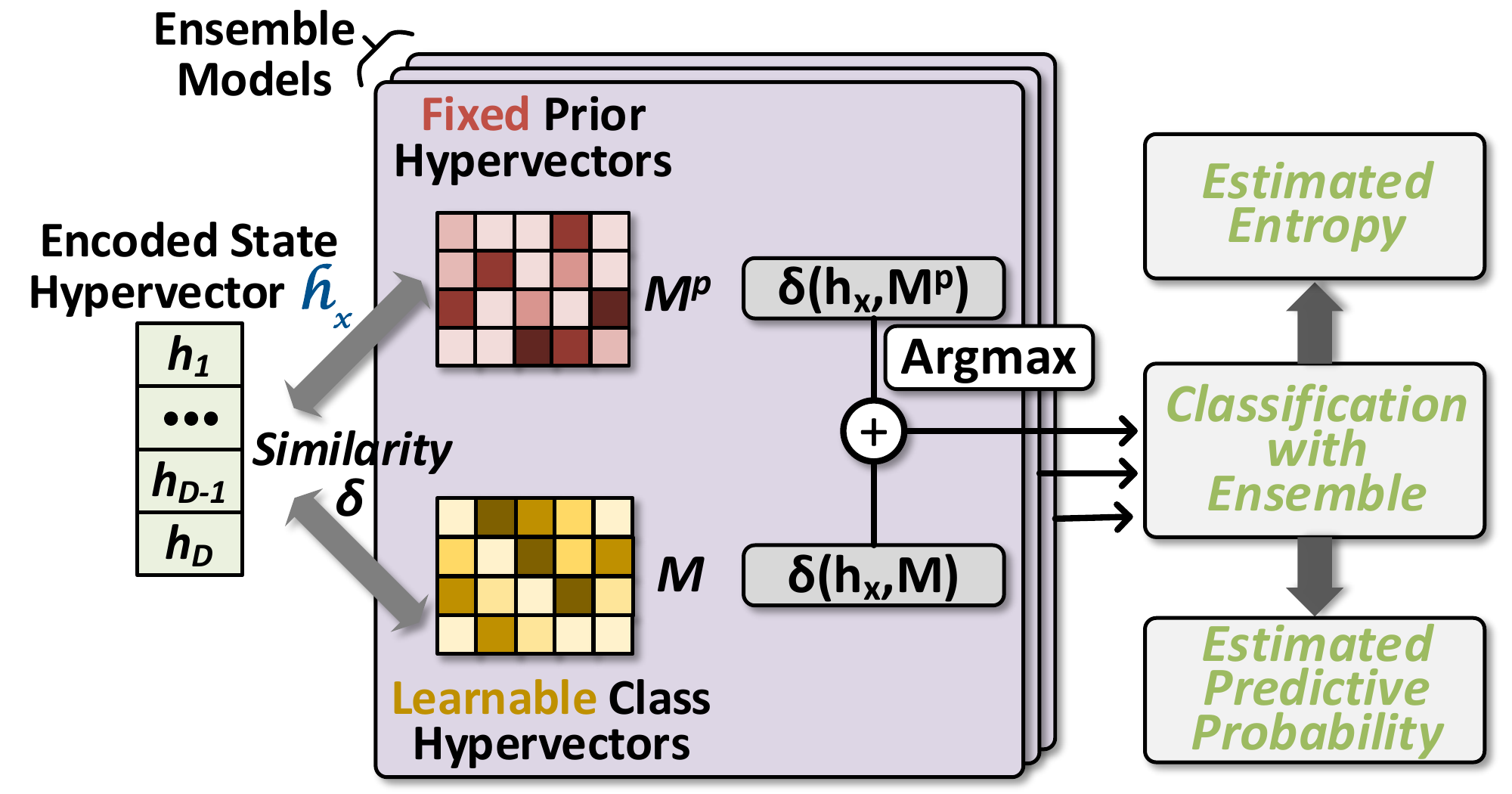}
    \caption{Ensemble HDC classification with prior hypervectors.}
    \label{fig:prior_ensemble}
\end{figure}

However, in \Design, we enhance the naive design in Section~\ref{sec:naive_ensemble} with HDC prior hypervectors that serve two purposes: \underline{first}, they are non-zero initializations of HDC models; \underline{second}, these hypervectors, as a whole, represent the prior model distribution or prior knowledge about a certain task. In Fig.~\ref{fig:prior_ensemble}, we give an overview of introducing prior hypervectors in HDC ensemble learning. We randomly sample these hypervectors with i.i.d. elements from the standard Gaussian distribution $\mathcal{N}(0,1)$, and as a way of model initialization, we enable prior hypervectors for every sub-model in the HDC ensemble. We refer to them as $\{\bm{M}_{1}^{p},\bm{M}_{2}^{p},\dots,\bm{M}_{E}^{p}\}$, in contrast to regular HDC models $\bm{M}$.

As shown in Fig.~\ref{fig:prior_ensemble}, we perform two separate similarity checks in each sub-model; one is between the encoded data $\bm{h}_{\bm{x}}$ and HDC prior hypervectors $\bm{M}^{p}$ and the other one is with model hypervectors $\bm{M}$. To make predictions, we combine these two sets of similarity values, e.g., for the first sub-model:
\begin{equation}
\label{eq:pred_prior}
    l_{pred}=\argmax_{\bm{m}_{l}\in\bm{M}_1,\bm{m}_{l}^{p}\in\bm{M}_{1}^{p}}\left[\delta\brackets{\bm{h}_{\bm{x}},\bm{m}_{l}}+\delta\brackets{\bm{h}_{\bm{x}},\bm{m}_{l}^{p}}\right]
\end{equation}
During the HDC model training, the prior hypervectors will not be updated and stay static and the model hypervectors follow slightly different update rules based on previous equations~\ref{eq:correct_pred} and~\ref{eq:wrong_pred}:
\begin{equation}
\label{eq:correct_pred_prior}
    \bm{m}_{l_{true}}=\bm{m}_{l_{true}}\oplus\lambda\brackets{1-S\brackets{\bm{h}_{\bm{x}},\bm{m}_{l_{true}}}}\bm{h}_{\bm{x}}
\end{equation}
\begin{equation}
\label{eq:wrong_pred_prior}
    \bm{m}_{l_{pred}}=\bm{m}_{l_{pred}}\oplus\lambda\brackets{S\brackets{\bm{h}_{\bm{x}},\bm{m}_{l_{pred}}}-1}\bm{h}_{\bm{x}}
\end{equation}
where $S\brackets{\bm{h}_{\bm{x}},\bm{m}_{l}}=\delta\brackets{\bm{h}_{\bm{x}},\bm{m}_{l}}+\delta\brackets{\bm{h}_{\bm{x}},\bm{m}_{l}^{p}}$ is the sum of two similarity values. Introducing prior hypervectors to training and inference helps ensemble sub-models learn with diverse viewing angles and allows them to disagree with each other when uncertain.

We will name $\bm{M}^{p}$ in Fig.~\ref{fig:prior_ensemble} as isolated prior hypervectors, meaning their similarity with queries is computed separately. However, there is an obvious alternative to equation~\ref{eq:pred_prior}: $l_{pred}=\argmax_{l}\delta\brackets{\bm{h}_{\bm{x}},\bm{m}_{l}+\bm{m}_{l}^{p}}$, where the prior hypervectors are combined/bundled with model hypervectors before calculating the similarity. We compare the two different ways below:
\begin{equation}
\label{eq:combine_prior}
    \delta\brackets{\bm{h}_{\bm{x}},\bm{m}_{l}\oplus\bm{m}_{l}^{p}}=\frac{real\brackets{\bm{h}_{\bm{x}}\cdot\brackets{\bm{m}_{l}\oplus\bm{m}_{l}^{p}}^{\dag}}}{||\bm{h}_{\bm{x}}||*||\bm{m}_{l}\oplus\bm{m}_{l}^{p}||}
\end{equation}
\begin{equation}
\label{eq:isolate_prior}
\begin{split}
    S\brackets{\bm{h}_{\bm{x}},\bm{m}_{l}}&=\delta\brackets{\bm{h}_{\bm{x}},\bm{m}_{l}}+\delta\brackets{\bm{h}_{\bm{x}},\bm{m}_{l}^{p}}\\
    &=\frac{real\brackets{\bm{h}_{\bm{x}}\cdot\bm{m}_{l}^{\dag}}}{||\bm{h}_{\bm{x}}||*||\bm{m}_{l}||}+\frac{real\brackets{\bm{h}_{\bm{x}}\cdot{\bm{m}_{l}^{p}}^{\dag}}}{||\bm{h}_{\bm{x}}||*||\bm{m}_{l}^{p}||}
\end{split}
\end{equation}

\begin{figure}[t]
    \centering
    \includegraphics[width=1\columnwidth]{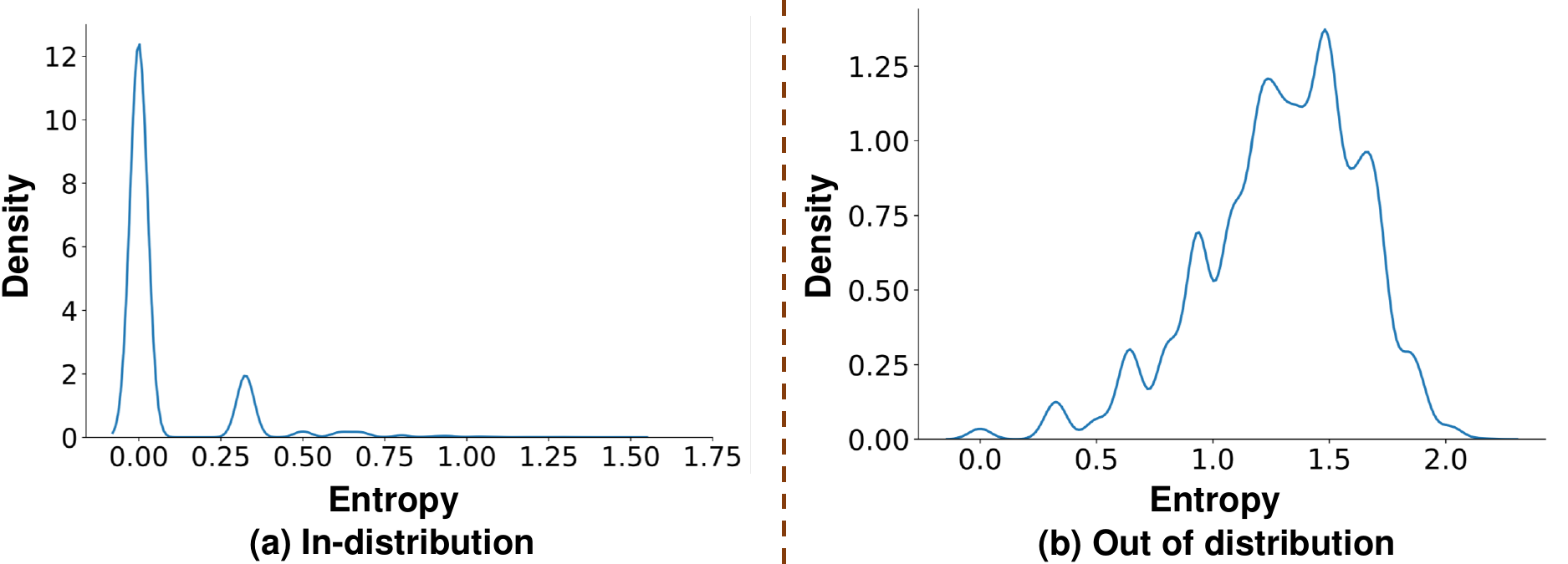}
    \caption{Histogram of predictive entropy for ensemble HDC classification with prior hypervectors on (a): in-distribution testing set and (b): OOD testing set.}
    \label{fig:hdc_prior}
\end{figure}

Fig.~\ref{fig:hdc_prior} compares the quality of uncertainty estimation between using isolated prior hypervectors and combining the prior with trainable models. With isolated prior hypervectors, in most cases, the HDC ensemble is able to reveal high uncertainty when predicting OOD test samples, showing an entropy value significantly larger than zero. However, when the prior hypervectors are combined into the model, the quality of uncertainty estimation is notably worse. The degradation may be attributed to the model hypervectors dominating the similarity computation. From equation~\ref{eq:combine_prior} and~\ref{eq:isolate_prior}, we observe that the difference is in the denominators: in the combined case, the norm is taken for the bundled hypervectors, whereas in the isolated case, the normalization happens separately, maintaining the influence of prior hypervectors on the overall similarity computation. More specifically, during HDC training, the model easily becomes much larger in norm ($||\bm{m}_{l}||\gg||\bm{m}_{l}^{p}||$), diminishing the effect of $\bm{m}_{l}^{p}$ and discouraging the diversity in sub-models.

In Algorithm~\ref{alg:UE_alg}, we present the pseudo-code for HDC ensemble learning with prior hypervectors, which is used in \Design to aid the estimation of model uncertainty.

\begin{algorithm}[t]
\caption{HDC Ensemble Learning with Prior}\label{alg:UE_alg}
\begin{algorithmic}
\State Assume an encoded training dataset with labels $\bcs{D}_{tr}$
\State Assume an HDC encoding matrix $\bm{\Theta}$
\State Assume $E$ HDC sub-models (model and prior): \\
\quad$\{\bm{M}_1,\bm{M}_2,\dots,\bm{M}_E\}$ and $\{\bm{M}_{1}^{p},\bm{M}_{2}^{p},\dots,\bm{M}_{E}^{p}\}$
\For{sub-model $\bm{M}_i$ and $\bm{M}_{i}^{p}$, $i\in[1,\dots,E]$ }
\State Pre-compute the similarity with prior $\bm{M}_{i}^{p}$ in eq.\ref{eq:isolate_prior}
\For{iteration $j$}
\For{$\{\bm{h}_{\bm{x}},l_{true}\}\in\bcs{D}_{tr}$}
\State Compute the similarity $S\brackets{\cdot}$ with eq.\ref{eq:isolate_prior}
\State Predict the label $l_{pred}$ using eq.\ref{eq:pred_prior}
\State Update $\bm{M}_{i}$ using eq.\ref{eq:correct_pred_prior} and eq.\ref{eq:wrong_pred_prior}
\EndFor
\EndFor
\EndFor
\end{algorithmic}
\end{algorithm}

\subsection{Computation Reuse and Neural Regeneration} \label{sec:efficient_design}

In this section, we explore the opportunities for further improving the efficiency of HDC uncertainty estimation in \Design. We will approach this via techniques such as a shared HDC encoder, encoded data reuse, similarity computation reuse, pre-normalization, and dynamic dimension regeneration.

\noindent\textbf{Reuse Computation in HDC Ensemble: }The key in ensemble learning is to construct multiple different sub-models; however, learning multiple models independently on every data point also leads to a several times higher cost. One difference between HDC-based and DNN-based ML models is that the HDC encoder is usually not updated once initialized. It is usually designed in advance since its main functionality is to generate high-dimensional and holographic representations for input data. With the analysis in previous sections, we also noticed that model (prior) hypervectors play a more important role in uncertainty estimation. Therefore, for HDC ensemble learning in \Design, we propose to share the HDC encoder among all sub-models, meaning that they will use a uniform high-dimensional representation. There are at least two immediate benefits of sharing the encoder: (1) In AL, we can encode the training data pool $\bcs{D}_{pool}$ only once in advance and reuse the encoded hypervectors $\bm{h}_{\bm{x}}$ for ensemble training and AL acquisition. They will significant amount of computation since both processes happen iteratively during the AL process. (2) Notice that both the HDC encoder and prior hypervectors are static, so, we can pre-compute and reuse their similarity values $real\brackets{\bm{h}_{\bm{x}}\cdot{\bm{m}_{l}^{p}}^{\dag}}/\brackets{||\bm{h}_{\bm{x}}||*||\bm{m}_{l}^{p}||}$ in equation~\ref{eq:isolate_prior}. In practice, we can also pre-compute the normalization term $||\bm{h}_{\bm{x}}||$ of the similarity value.

\noindent\textbf{HDC Neural Regeneration for Better Acquisition Efficiency: }
As operations in \Design are centered around hypervectors, the computation cost scales with dimensionality. Thus, cutting down the dimensionality becomes a natural choice to further reduce the runtime cost of \Design acquisition. We propose to leverage NeuralHD~\cite{zou2021scalable} to obtain even more lightweight HDC classification models. It introduces a mechanism to dynamically update the HDC encoder and is compatible with existing HDC classification algorithms including \Design. Motivated by the regenerative ability of the brain, after a few epochs of HDC model update, NueralHD evaluates the significance of each dimension to the classification accuracy. The metric used here is the normalized variance across classes at each dimension of class hypervectors.
NeuralHD updates the positional hypervectors $\rho$ of the HDC encoder by regenerating the dimensions that show low variance in class hypervectors. More specifically, we sample new values for those dimensions from standard Gaussian distribution again as in Section~\ref{sec:HDC_encoding}. This process makes NeuralHD more efficient in the utilization of hyperdimensions and maintains high accuracy with much smaller HDC models (i.e., much fewer hyperdimensions). Due to the limited space, for more details regarding this algorithm, we refer readers to the original paper~\cite{zou2021scalable}. For our design, with NeuralHD regeneration, we reduce the hypervector dimensionality in \Design by up to 50\%.

\section{HDC-based Active Learning} \label{sec:AL}

\begin{figure}[t]
    \centering
    \includegraphics[width=1\columnwidth]{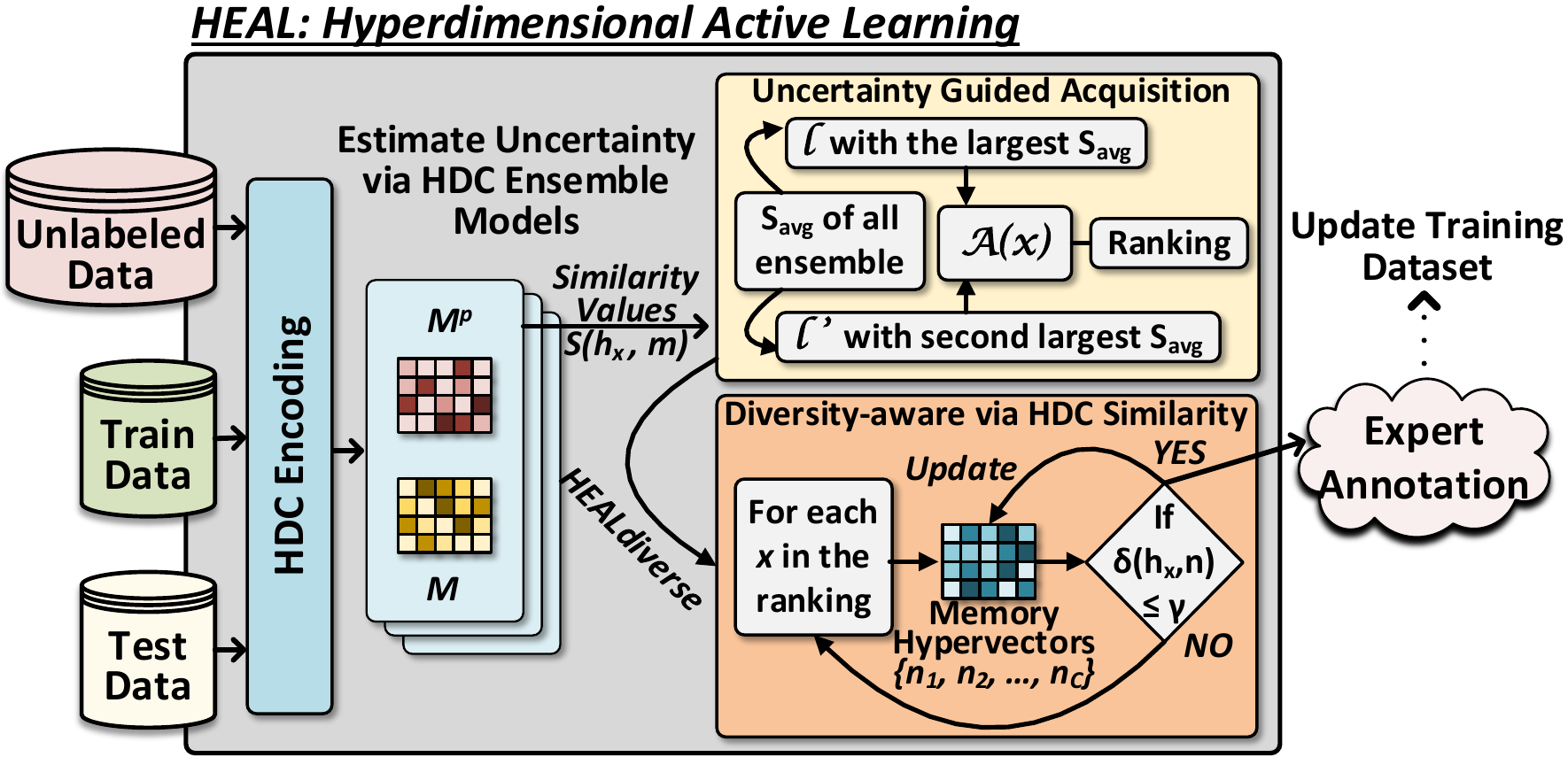}
    \caption{Overview of our proposed AL framework comprised of uncertainty-guided acquisition and diversity-aware acquisition.}
    \label{fig:HEAL_overview}
\end{figure}

In Fig.~\ref{fig:HEAL_overview}, we present the overview of \Design, our proposed HDC-based AL framework. It has three main components: HDC ensemble learning for uncertainty estimation (covered in the previous Section~\ref{sec:UE}), \Design uncertainty-based acquisition, and HDC diversity-enhanced AL. In this section, we will focus on the latter two.

\subsection{Uncertainty-based Acquisition via Hypervector Similarity}

It is common for AL techniques to leverage probabilities in the acquisition function because DNNs usually add a softmax layer to normalize logits and get probability values in the range of 0 to 1. However, HDC does not give outputs interpretable as probabilities nor does it learn based on cross-entropy loss. This means that adding an extra softmax layer will not give informative predictive probabilities. 
Instead of using softmax, in equation~\ref{eq:estimate_prob}, we estimate the predictive probability from predictions of ensemble sub-models. Although we have shown that it is quite useful for qualitative analysis, this is inevitably a coarse approximation and will limit its usage in uncertainty-based AL. 

With no access to predictive probabilities, prior AL techniques like confidence and entropy sampling are unsuitable for HDC. However, we noticed that the margin sampling, although based on probabilities in the original design, can be adapted for HDC. It captures model uncertainty through the margin between the top two predictions, which generalizes beyond probability values. 
In \Design, we propose to directly utilize the similarity values from equation~\ref{eq:isolate_prior} and compute the prediction margin in the AL acquisition function:
\begin{equation}
\label{eq:marg_sampling}
    \bcs{A}\brackets{\bm{x}}=S_{avg}\brackets{\bm{h}_{\bm{x}},\bm{m}_{l'}}-S_{avg}\brackets{\bm{h}_{\bm{x}},\bm{m}_{l}}
\end{equation}
The value of $\bcs{A}\brackets{\bm{x}}$ becomes larger when the similarity with $\bm{m}_{l'}$ gets closer to the one with $\bm{m}_{l}$, i.e., higher uncertainty about $\bm{x}$.
We apply ensemble model averaging for similarity values; $l$ is the predicted label with the highest average similarity values and $l'$ is the one with the second highest values.
\begin{equation}
\label{eq:model_avg}
    S_{avg}\brackets{\bm{h}_{\bm{x}},\bm{m}_{l}}=\frac{\sum_{i=1}^{E}S\brackets{\bm{h}_{\bm{x}},\bm{m}_{l}^{i}}}{E}
\end{equation}
where $\bm{m}_{l}^{i}$ represents the model(prior) hypervectors of class $l$ corresponding to the $i$-th sub-model. In every acquisition step, we sample a batch of data points from $\bcs{D}_{pool}$, which rank top-$K$ in the values of $\bcs{A}\brackets{x}$.

HDC ensemble with prior hypervectors plays a key role in the \Design acquisition process. After training, each sub-model can be confident about its prediction, giving high similarity values for the predicted class. However, sub-models will predict different labels for data points that the ensemble model as a whole is uncertain about. With ensemble model averaging in equation~\ref{eq:model_avg}, the similarity value for a particular class is much lower due to the disagreement, and the margin sampling in equation~\ref{eq:marg_sampling} will capture the narrowed gap between top-2 predictions.

\subsection{Diversity Metric via Hypervector Memorization}
\label{sec:diversity_AL}

One common challenge in uncertainty-based AL is that the algorithm tends to select similar samples in batch acquisition. Notice that during batch acquisition, the model is not updated immediately until a full batch of data points is annotated. Therefore, a top-$K$ ranking will repeatedly select helpful data points that contain duplicate information. As we mentioned in Section~\ref{sec:challenges}, DNN-based AL methods cope with this problem via joint mutual information, representative sampling, and data mining algorithms, although with significant acquisition overheads due to these added components.

In this section, we propose an efficient diversity metric that helps \Design acquire not only informative but also diverse data points, without introducing costly computations as in prior DNN-based methods. 
We utilize lightweight HDC operations and leverage the intrinsic memorization capability of hypervectors to achieve diverse acquisition.
We notice that the requirement for diversity can be achieved by checking the similarity between the candidate data point and existing points in the current batch. In other words, hypervector similarity checks can be a strong tool for filtering out duplicate data points. This diversity metric can be seamlessly included in \Design since the dataset has already been encoded to hypervector representations. 

As shown in Fig.~\ref{fig:HEAL_overview}, instead of computing pair-wise similarity, we consider the acquired data points altogether by constructing a memorization hypervector for each class $\{\bm{n}_{1},\bm{n}_{2},\dots,\bm{n}_{C}\}$. 
These hypervectors memorize and categorize the acquired data points according to the pseudo label $\hat{l}$, i.e., the label predicted by all HDC sub-models via voting. They are initialized with all-zero vectors.
To begin with, we perform inference on the unlabeled dataset $\bcs{D}_{pool}$ and rank it via the acquisition function $\bcs{A}\brackets{\bm{x}}$ to prioritize the data points for which the model has lower confidence.
The first acquired data point $\bm{x}_{1}$ (i.e., the one with the largest value of $\bcs{A}\brackets{\bm{x}}$) will be automatically added to the batch. We update the memory hypervector that corresponds to the pseudo label: $\bm{n}_{\hat{l}_{\bm{x}_1}}=\bm{n}_{\hat{l}_{\bm{x}_1}}\oplus\bm{h}_{\bm{x}_1}$. For the next candidate data point $\bm{x}_2$ that ranked second, if it has the same pseudo label as $\bm{x}_1$, we will check its similarity with the memorization hypervector: $\delta\brackets{\bm{h}_{\bm{x}_2},\bm{n}_{\hat{l}_{\bm{x}_1}}}$. We only acquire this sample when it shows low similarity values and discard it if otherwise. In our implementation, we use a similarity threshold $\gamma=0.4$. If its prediction leads to a yet empty memory hypervector, it will be acquired directly. We will repeat this process until the batch has been filled with acquired data points. For the next step of batch acquisition, the memory hypervectors are re-initialized. 

We present the pseudo-code for \Design in Algorithm~\ref{alg:AL_alg}. Before the acquisition process, we first pre-encode $\bcs{D}_{pool}$ and pre-compute similarities with prior hypervectors as mentioned in Section~\ref{sec:efficient_design}. Then in each acquisition step $t$, we annotate the acquired data points $\bcs{B}$ to the training dataset, remove them from $\bcs{D}_{pool}$, and then train the ensemble model on the new training set.

\begin{algorithm}[t]
\caption{\Design: HDC Uncertainty \& Diversity-aware AL}\label{alg:AL_alg}
\begin{algorithmic}
\State Assume $\bcs{D}_{pool}$, $n_{init}$, $K$, $\gamma$, and $t=1$
\State Initialize the training dataset $\bcs{D}_{tr}^{0}$ with $n_{init}$ points
\State Initialize HDC encoding matrix $\bm{\Theta}\sim\{\mathcal{N}\brackets{0,1}\}^{n\times D}$
\State Initialize $E$ HDC sub-models (model and prior): \\
\quad$\bm{M}\sim\{0\}^{C\times D}$ and $\bm{M}^{p}\sim\{\mathcal{N}(0,1)\}^{C\times D}$
\State Encode $\bcs{D}_{pool}$ to $\bcs{D}_{pool\_en}$ using eq.\ref{eq:hdc_encode}
\For{$\bm{h}_{\bm{x}}\in\bcs{D}_{pool\_en}$}
\For{$\bm{M}_{i}^{p}$, $i\in[1,\dots,E]$}
\State Pre-compute similarities with prior $\bm{M}_{i}^{p}$ in eq.\ref{eq:isolate_prior}
\EndFor
\EndFor
\For{Acquisition step t}
\State $b\leftarrow0$, $\bcs{B}\leftarrow\emptyset$
\State Train HDC ensemble model on $\bcs{D}_{tr}^{t}$ (Alg.~\ref{alg:UE_alg})
\For{$\bm{h}_{\bm{x}}\in\bcs{D}_{pool\_en}^{t}$}
\For{$\bm{M}_{i}$, $i\in[1,\dots,E]$}
\State Compute the similarity $S\brackets{\cdot}$ with eq.\ref{eq:isolate_prior} 
\EndFor
\State Evaluate with acquisition function $\bcs{A}\brackets{\bm{x}}$ in eq.~\ref{eq:marg_sampling}
\State Annotate the $\bm{x}$ with pseudo label $\hat{l_{\bm{x}}}$
\EndFor
\State Rank $\bcs{D}_{pool\_en}^{t}$ according to $\bcs{A}\brackets{\bm{x}}$ in descending order
\State Initialize $\bm{n}_{1},\bm{n}_{1},\dots,\bm{n}_{C}\sim\{0\}^{D}$
\For{$\{\bm{h}_{\bm{x}},\hat{l_{\bm{x}}}\}\in\bcs{D}_{pool\_en}^{t}$}
\If{$\delta\brackets{\bm{h}_{\bm{x}},\bm{n}_{\hat{l_{\bm{x}}}}}\leq\gamma$}
\State $\bcs{B}\leftarrow\bcs{B}\cup\{\bm{h}_{\bm{x}},\hat{l_{\bm{x}}}\}$
\State$\bm{n}_{\hat{l_{\bm{x}}}}\leftarrow\bm{n}_{\hat{l_{\bm{x}}}}\oplus\bm{h}_{\bm{x}}$
\State $b\leftarrow b+1$
\EndIf
\If{$b \geq K$}
\State \textbf{break}
\EndIf
\EndFor
\State Annotate $\bcs{B}$ with true labels $l_{true}$
\State $\bcs{D}_{tr}^{t}\leftarrow\bcs{D}_{tr}^{t-1}\cup\bcs{B}$
\State $\bcs{D}_{pool\_en}^{t}\leftarrow\bcs{D}_{pool\_en}^{t-1}\setminus\bcs{B}$
\State $t\leftarrow t+1$
\EndFor
\end{algorithmic}
\end{algorithm}

\begin{table}[b]
\caption{Details of the datasets used for experiments}
\centerline{
\resizebox{1.0\columnwidth}{!}{
\begin{tabular}{c|c|c|c|c}
\toprule
\textbf{Datasets}        & \textbf{\# Train Samples} & \textbf{\# Test Samples} & \textbf{\# Features} & \textbf{\# Classes}   \\
\midrule
\midrule

\textbf{ISOLET~\cite{Dua:2019}}     & 5847 & 1950   & 617         & 26\\
\midrule
\textbf{UCIHAR~\cite{anguita2013public}}     & 5825 & 1942    & 561         & 12  \\
\midrule
\textbf{DSADS~\cite{barshan2014recognizing}}     & 6840 & 2280    & 405         & 19  \\
\midrule
\textbf{PAMAP~\cite{reiss2012introducing}}     & 5484 & 1828   & 243          & 19    \\
\bottomrule
\end{tabular}
}} \label{tbl:dataset}
\end{table}

\begin{figure*}[t]
    \centering
    \includegraphics[width=2\columnwidth]{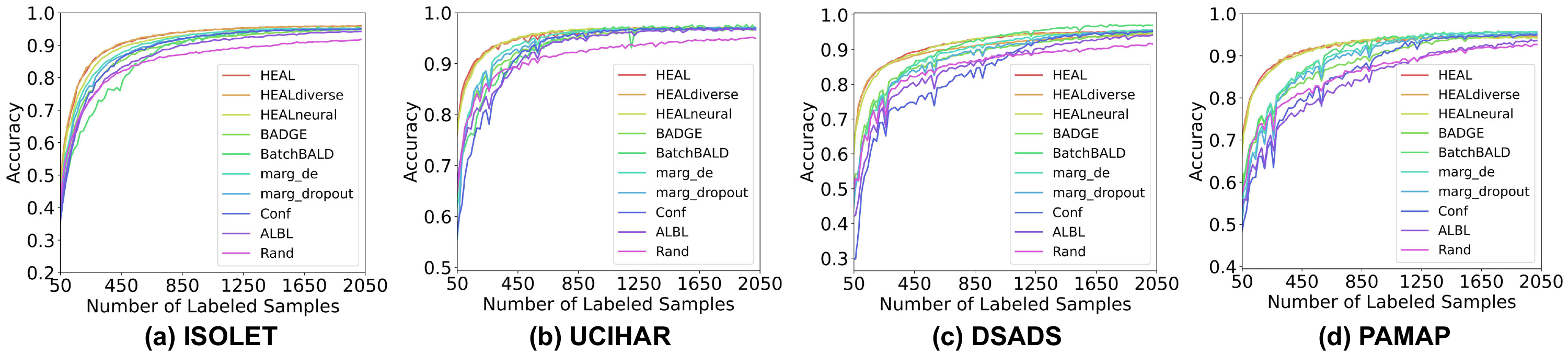}
    \caption{Average learning curves for different AL algorithms on four datasets. The initial labeled training dataset has 20 samples and the AL batch size is 20.}
    \label{fig:no_dup_learning_curve}
\end{figure*}

\section{Experiments} \label{sec:Ex}

\subsection{Experimental Settings} \label{sec:ex_settings}
To evaluate the performance of \Design, we compare it against several existing AL algorithms that are widely applied, including traditional methods that are based on simple measures, and modern methods that are either non-Bayesian or Bayesian in terms of how uncertainty is estimated. Our baselines also include AL algorithms that explicitly consider diversity in acquisition. The following is a list of the baseline algorithms used in comparison.

\begin{figure*}[t]
    \centering
    \includegraphics[width=2\columnwidth]{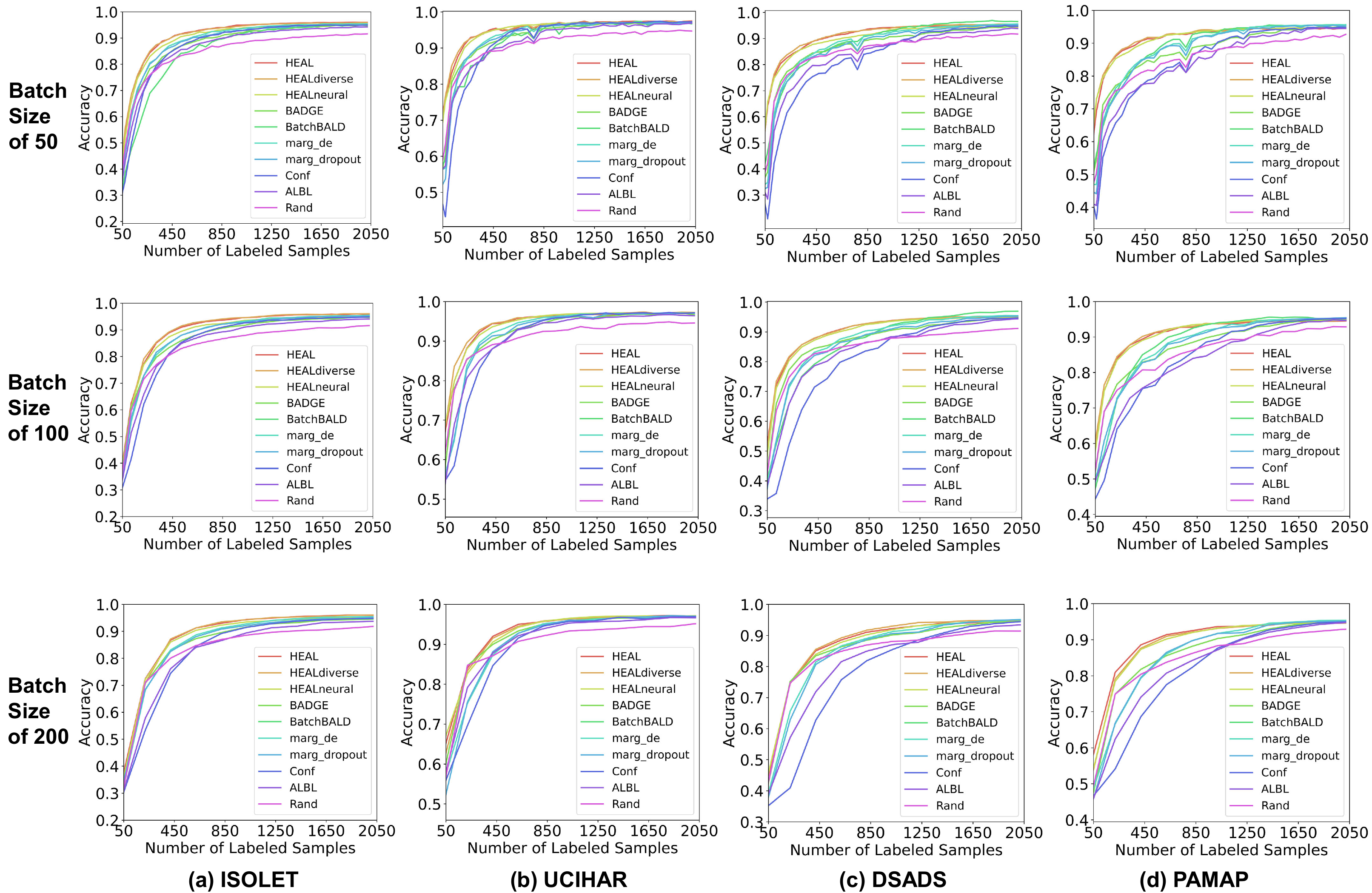}
    \caption{Average learning curves for different AL acquisition batch sizes. Each column (row) corresponds to a specific dataset (AL batch size).}
    \label{fig:no_dup_batch_size}
\end{figure*}

\begin{figure*}[t]
    \centering
    \includegraphics[width=2\columnwidth]{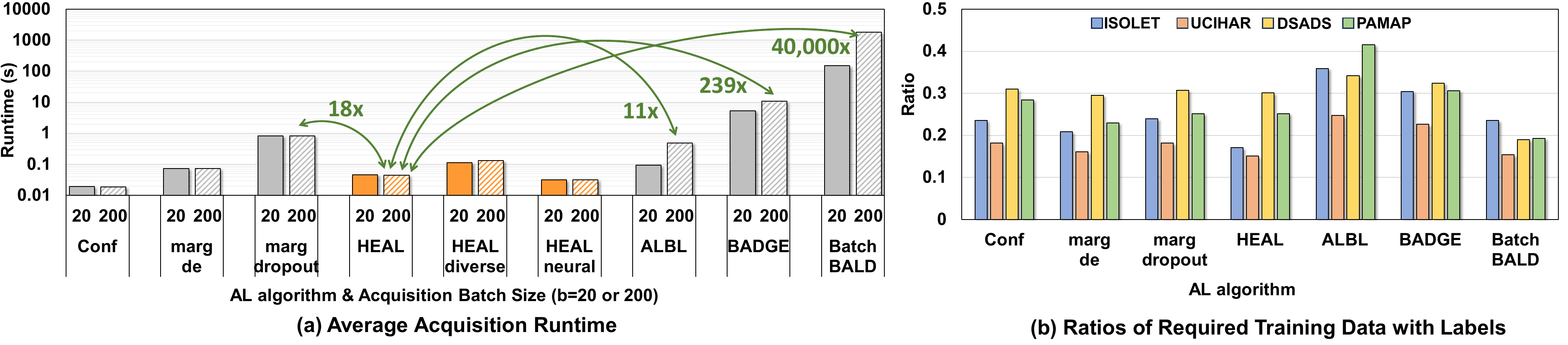}
    \caption{(a) Active learning acquisition runtime per batch for different AL algorithms and batch sizes. The runtime is averaged over four datasets. (b) At $b=20$, the ratios of required labeled data for different AL methods.}
    \label{fig:no_dup_query_runtime}
\end{figure*}

\begin{enumerate}[leftmargin=*]
\item \textbf{Rand:} A naive design (i.e., non-AL) that randomly acquires unlabeled samples for expert annotation.
\item \textbf{Conf (Confidence Sampling):} uses simple model confidence as the sign for uncertainty~\cite{wang2014new}. The acquisition starts with the sample with the lowest predicted class probability.
\item \textbf{Marg (Margin Sampling):} An uncertainty-based algorithm, whose uncertainty metric is built upon the probability difference between the top-2 predicted classes~\cite{roth2006margin}. The sample with the smallest margin will be selected first for annotation. Margin sampling can be enhanced by BNNs for better uncertainty estimation. 
    \begin{enumerate}
    \item Marg\_de: It uses Deep Ensemble as the BNN backbone.
    \item Marg\_dropout: uses MC-Dropout as the BNN backbone.
    \end{enumerate}
\item \textbf{ALBL:} A hybrid AL method that aims at the balance between uncertainty-based (Conf) and diversity-based (Coreset) AL algorithms via bandit-style selection~\cite{hsu2015active}.
\item \textbf{BADGE:} An intelligent hybrid method that uses gradient information in the DNN classifier to incorporate both predictive uncertainty and sample diversity into acquisition~\cite{ash2019deep}.
\item \textbf{BatchBALD:} An information-theoretic AL method that identifies the most informative samples based on estimated information gain, computed via Bayesian models~\cite{kirsch2019batchbald}. The metric used considers both model uncertainty and diversity of the acquired batch.
\end{enumerate}
As for our proposed \Design, since many of its components are optional during practical implementations, we evaluated it with the following different settings:
\begin{itemize}
    \item \Design: the bare bone of our uncertainty-based AL framework using HDC ensemble with prior hypervectors.
    \item \Designn: enhanced by hyperdimension regeneration for a more lightweight model and faster acquisition. 
    \item \Designd: enhanced by HDC memory hypervectors for diversity-aware batch acquisition.
\end{itemize}

\begin{figure*}[t]
    \centering
    \includegraphics[width=2\columnwidth]{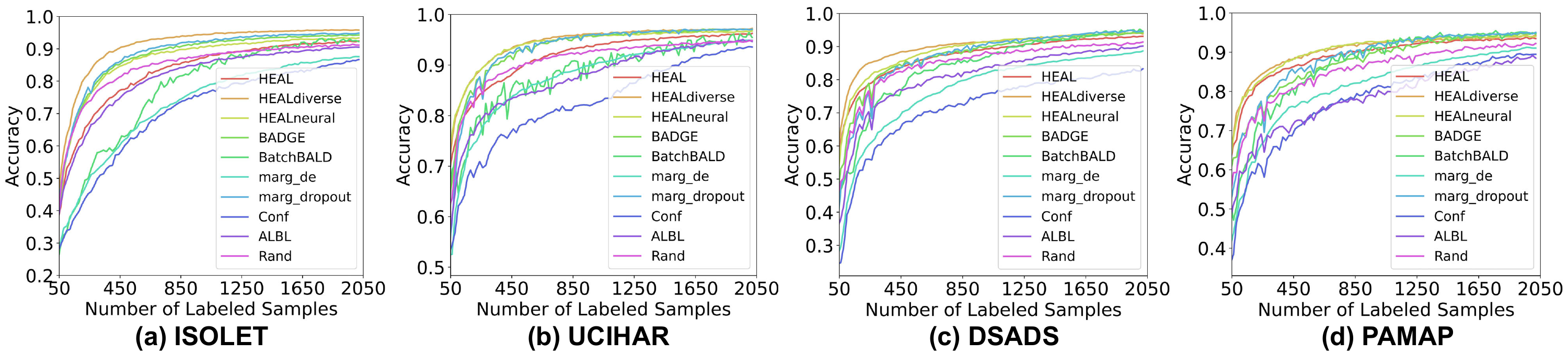}
    \caption{Average learning curves for different AL algorithms on four datasets with duplicate samples. The initial labeled training dataset has 20 samples and the AL batch size is 20.}
    \label{fig:with_dup_learning_curve}
\end{figure*}

\begin{figure*}[t]
    \centering
    \includegraphics[width=1.5\columnwidth]{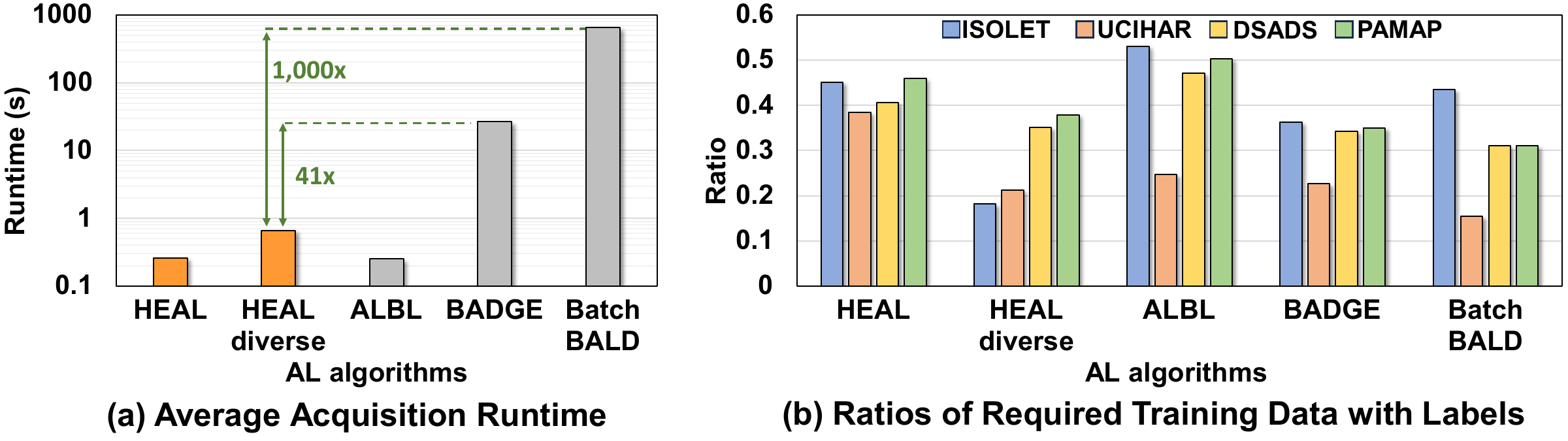}
    \caption{(a) Active learning acquisition runtime per batch for different AL algorithms. The runtime is averaged over four datasets. (b) The ratios of required labeled data for different AL methods. The batch size is set to $b=20$.}
    \label{fig:overhead_analysis}
\end{figure*}

Our AL evaluation follows the batch-mode setup; for all algorithms, it starts with 20 initial labeled training samples ($|\mathcal{D}_{pool}|=20$), and the acquisition batch size $b$ for each AL step ranges from 20 to 200. As for the model backbone, similar to BatchBALD, we select the multilayer perception (MLP) model for all DNN/BNN-based AL on all datasets; it has two hidden layers and each layer has 256 neurons. The models are trained on Pytorch using cross-entropy loss and the Adam optimizer. For \Design and \Designd, we use hypervectors with $D=2000$; and \Designn, as we mentioned before, uses $D=1000$. For all algorithms, we train the classifier from scratch at every step of acquisition until the training accuracy hits 99\%. All experiments are repeated five times and averaged.

We showcase our proposed AL frameworks on four different open datasets, as shown in Table~\ref{tbl:dataset}. The first dataset is a speech recognition dataset and the rest three are for human activity recognition tasks. In practice, human activity recognition mainly involves on-body multi-sensor data analysis, where collecting labeled data for diverse user groups takes a significant amount of effort. Our experiments aim to show the effectiveness of AL algorithms, especially \Design, on saving annotation costs in similar tasks. Notice that the DSADS and PAMAP datasets undergo a widely applied data preprocessing as described in~\cite{wang2018stratified}.

\subsection{\Design Active Learning Performance and Efficiency} \label{sec:HEAL_perf}

\begin{figure}[t]
    \centering
    \includegraphics[width=1\columnwidth]{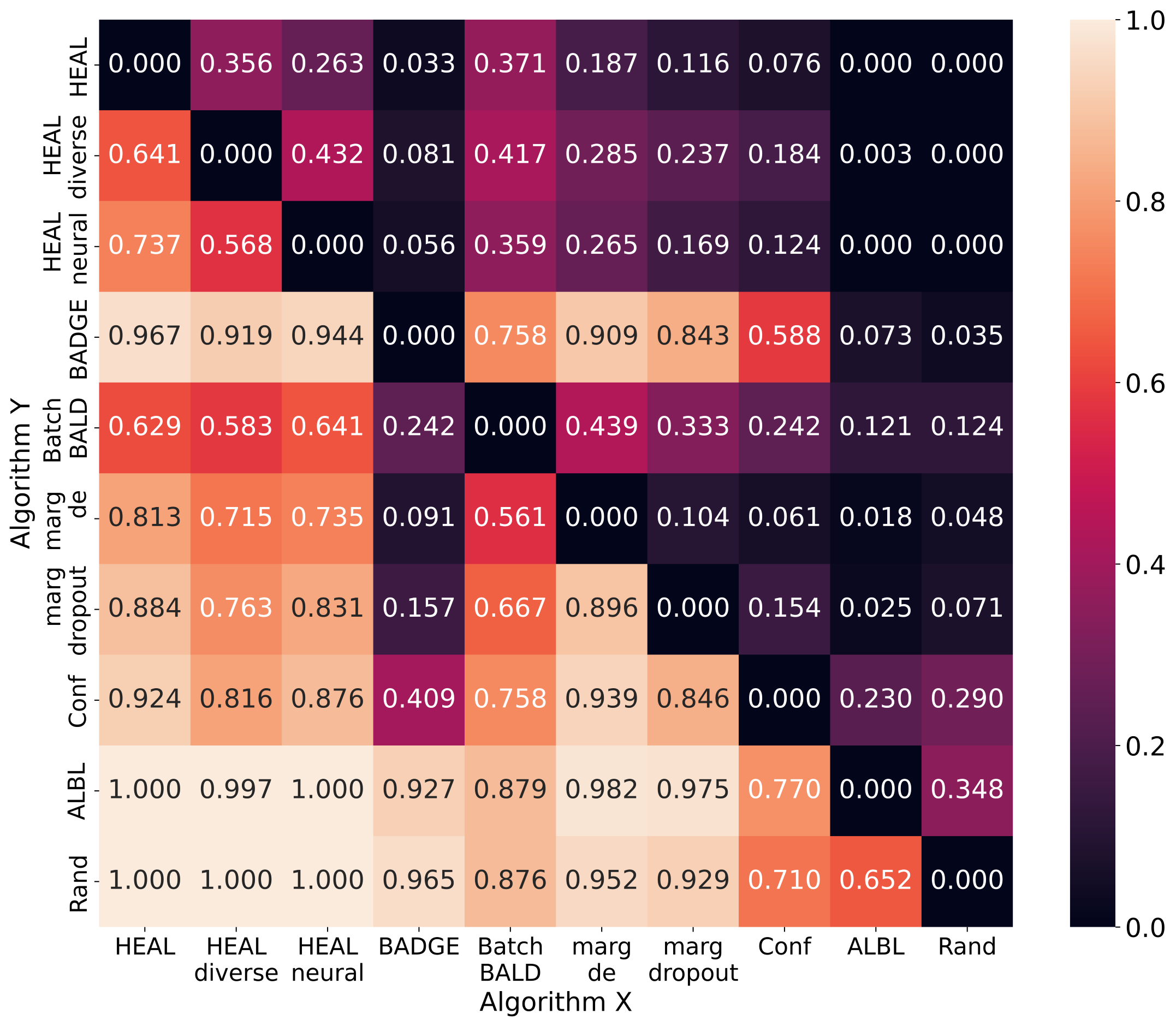}
    \caption{Pair-wise comparison averaged on datasets without duplicate samples}
    \label{fig:no_dup_matrix}
\end{figure}

\begin{figure}[t]
    \centering
    \includegraphics[width=1\columnwidth]{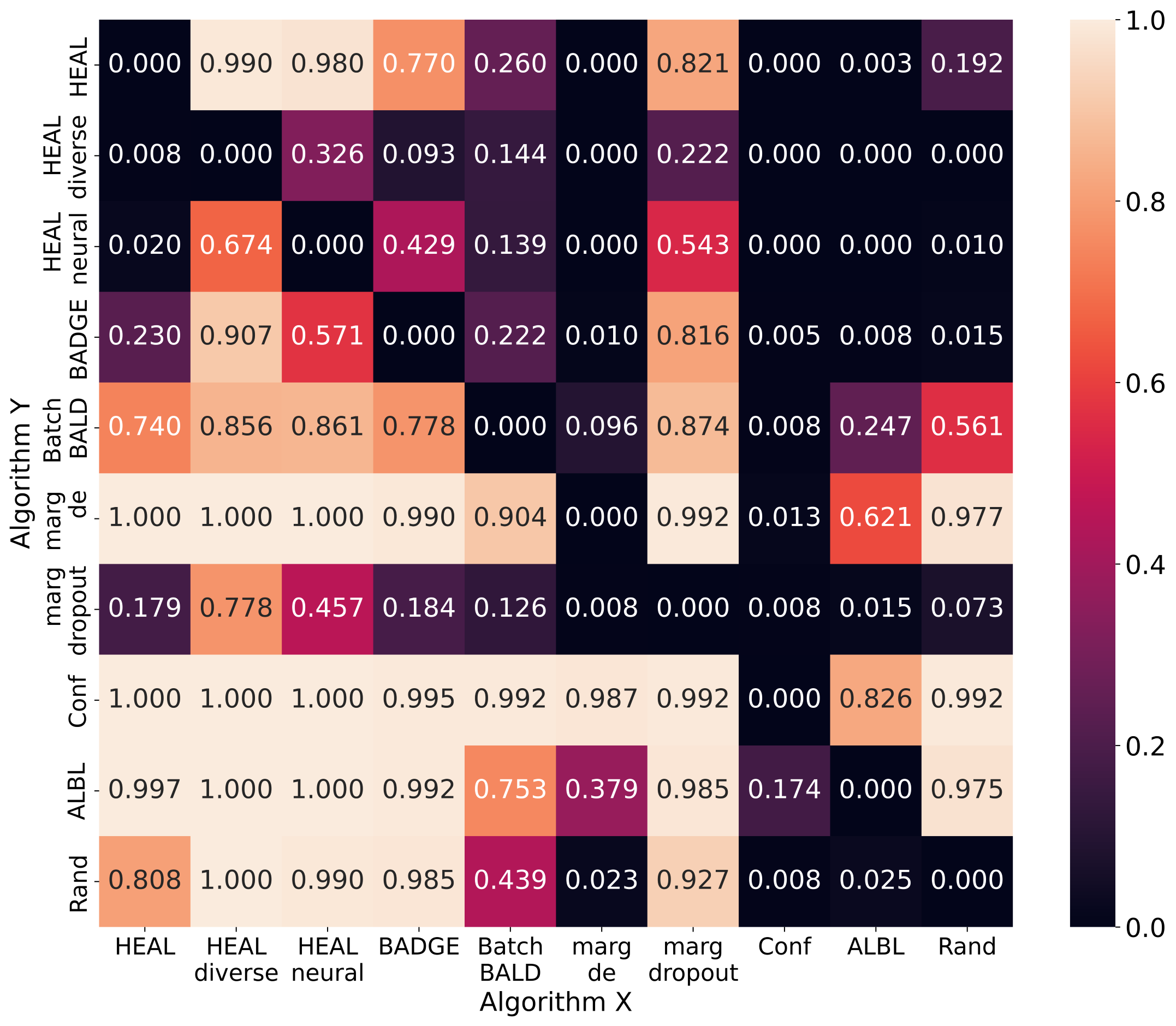}
    \caption{Pair-wise comparison averaged on datasets with duplicate samples}
    \label{fig:with_dup_matrix}
\end{figure}

The learning curve is an intuitive way to evaluate the effectiveness of AL algorithms, which record the testing accuracy at each step of acquisition with an increasing number of labeled samples. In Fig.~\ref{fig:no_dup_learning_curve}, we show the averaged testing accuracy of five runs for all four datasets and different AL algorithms. The batch size of acquisition is $b=20$. In Fig.~\ref{fig:no_dup_batch_size}, we present learning curves with other batch sizes ($b={50,100,200}$). Firstly, we observe that all AL algorithms eventually outperform random acquisition by a noticeable gap. Most AL algorithms help reach model convergence much faster than the case without AL. As for the performance of \Design methods with different configurations, they often achieve significantly higher testing accuracy than other baselines during the first half of the curve, i.e., with less than 1000 labeled samples. When closer to model convergence, \Design enhanced HDC classifier is comparable to hybrid AL methods including BADGE and BatchBALD. Albeit using similar acquisition metrics as margin sampling, \Design achieves better AL quality, thanks to the optimal combination of HDC's intrinsic learning efficiency and our uniquely developed uncertainty estimation approach for HDC-based classifiers. For more comparison between different variants of margin sampling and \Design, please refer to Fig.~\ref{fig:bayes_comparison}. For illustrations of HDC inherent learning efficiency and its comparison with \Design, please refer to Fig.~\ref{fig:HDC_HEAL}.

To illustrate the efficiency of our AL framework, we collect the acquisition runtime for most algorithms (including \Design) using Intel Core i7-12700 CPU; except for BatchBALD, which is not efficient and scalable on CPU platforms, we use NVIDIA RTX 4090 GPU instead. In Fig.~\ref{fig:no_dup_query_runtime}(a), we compare the average acquisition runtime of each AL method with two different batch size settings. As highlighted in this figure, \Design and its variants have notably faster acquisition compared to most baselines. When $b=200$, the speedups of \Design over marginal sampling with dropout, ALBL, BADGE, and BatchBALD are 18$\times$, 11$\times$, 239$\times$, and more than 40000$\times$, respectively. Confidence sampling is the fastest in acquisition due to its non-Bayesian uncertainty estimation. However, this naive estimation leads to its sub-optimal performance in AL. With the help of NeuralHD, \Designn is about 45\% faster than regular \Design in acquisition. Due to the diversity-aware acquisition module, there is a relatively small overhead in \Designd. In Section~\ref{sec:perf_dup}, we will present the benefits brought by this extra module. In Fig.~\ref{fig:no_dup_query_runtime} (b), we show the size of the labeled training dataset needed for each AL method to achieve 99\% of the accuracy obtained with the full dataset. This metric represents how much annotation effort can be saved with AL. The figure shows that \Design is on average better than most baselines except the information-theoretic BatchBALD which comes at a huge cost. Note that the acquisition in BatchBALD is significantly slower than others even if it is the only method run on a powerful GPU. This is because BatchBALD suffers from the combinatorial explosion in its estimation of joint distributions~\cite{kirsch2019batchbald}.

\subsection{AL in datasets with duplicated samples} \label{sec:perf_dup}

As we mentioned in Section~\ref{sec:diversity_AL}, datasets with a large number of similar samples pose challenges to many existing AL algorithms. Therefore, in this section, we ramp up the difficulty and evaluate these AL methods on specially modified datasets. For each of the previously tested datasets, we copy the training dataset four times, meaning that each unique sample now has five duplicates. We then repeat the evaluations in Section~\ref{sec:HEAL_perf} and record the learning curves in Fig.~\ref{fig:with_dup_learning_curve}. As expected, AL methods without effective diversity metrics such as confidence sampling and marg\_de suffer from significant degradation in AL performance. In addition, the performance of ALBL over confidence sampling hints at the benefit of filtering out duplicate samples during acquisition. Nevertheless, many AL algorithms are showing acquisition quality worse than random selection. In contrast, methods like \Designd and BADGE still maintain high acquisition efficiency with \Designd giving the highest testing accuracy in the first half of the learning curve. Methods including BADGE and BatchBALD are also among the best performing AL methods during the second half, however, their acquisition costs are orders of magnitude larger. Also interestingly, we observe that methods that rely solely on ensemble generally perform poorly, as their sub-models are prone to be similar due to training on an overflow of duplicated samples and thereby compromise their uncertainty estimation. Fig.~\ref{fig:overhead_analysis}(a) shows that \Designd is 41$\times$ (1000$\times$) faster than BADGE (BatchBALD) in terms of the average acquisition runtime. Fig.~\ref{fig:overhead_analysis}(b) shows that \Designd is among the best performing AL methods for all datasets and significantly outperforms the regular \Design design.

\subsection{Pair-wise comparison}

To better compare different AL methods comprehensively, we apply a pair-wise comparison method proposed in~\cite{ash2019deep}. Every time algorithm X beats algorithm Y in terms of testing accuracy, the latter accumulates a certain amount of penalty. We normalize the values to $[0,1]$. Fig.~\ref{fig:no_dup_matrix} is the pair-wise comparison averaged on datasets without duplicate samples, and Fig.~\ref{fig:with_dup_matrix} is averaged on datasets with duplicate samples. A better-performing AL method shows more small values (i.e., dark color) in a row, e.g., \Design outperforms in Fig.~\ref{fig:no_dup_matrix} and \Designd in Fig.~\ref{fig:with_dup_matrix}.

\subsection{Benefits of using advanced uncertainty estimation in AL}

\begin{figure}[t]
    \centering
    \includegraphics[width=0.6\columnwidth]{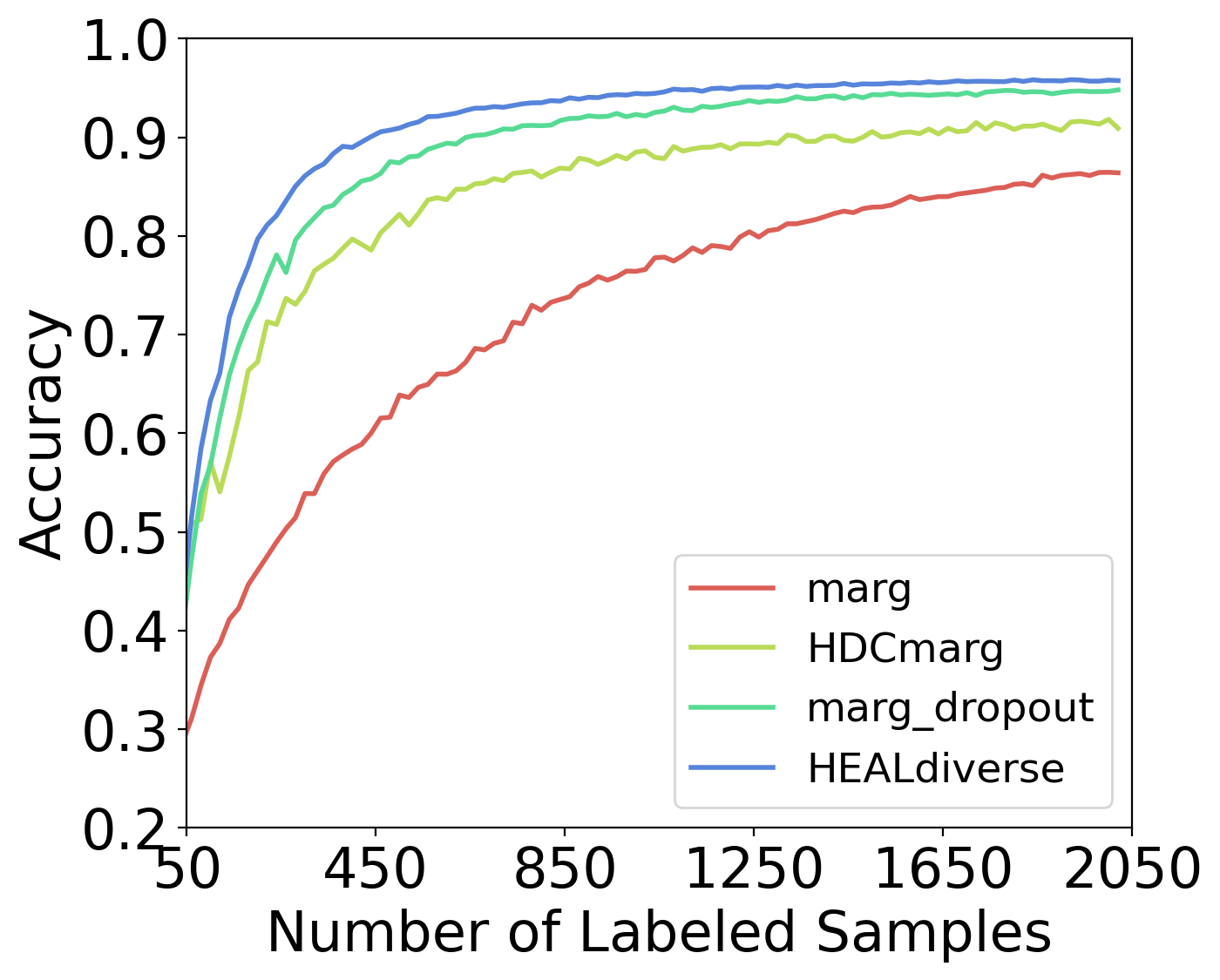}
    \caption{Learning curve comparison for AL methods with or without advanced uncertainty estimation on ISOLET dataset with duplicated samples}
    \label{fig:bayes_comparison}
\end{figure}

In this section, we illustrate the benefits of designing non-trivial uncertainty estimation techniques for AL by comparing methods with or without these techniques. In Fig.~\ref{fig:bayes_comparison}, 'marg' stands for the basic margin sampling without using any BNNs, and 'HDCmarg' refers to simple HDC similarity-based margin sampling without HDC ensemble models with prior hypervectors. In general, \Designd and margin sampling with MC-dropout significantly outperform their naive versions.

\subsection{Comparison against HDC classifiers without AL}

\begin{figure}[t]
    \centering
    \includegraphics[width=0.6\columnwidth]{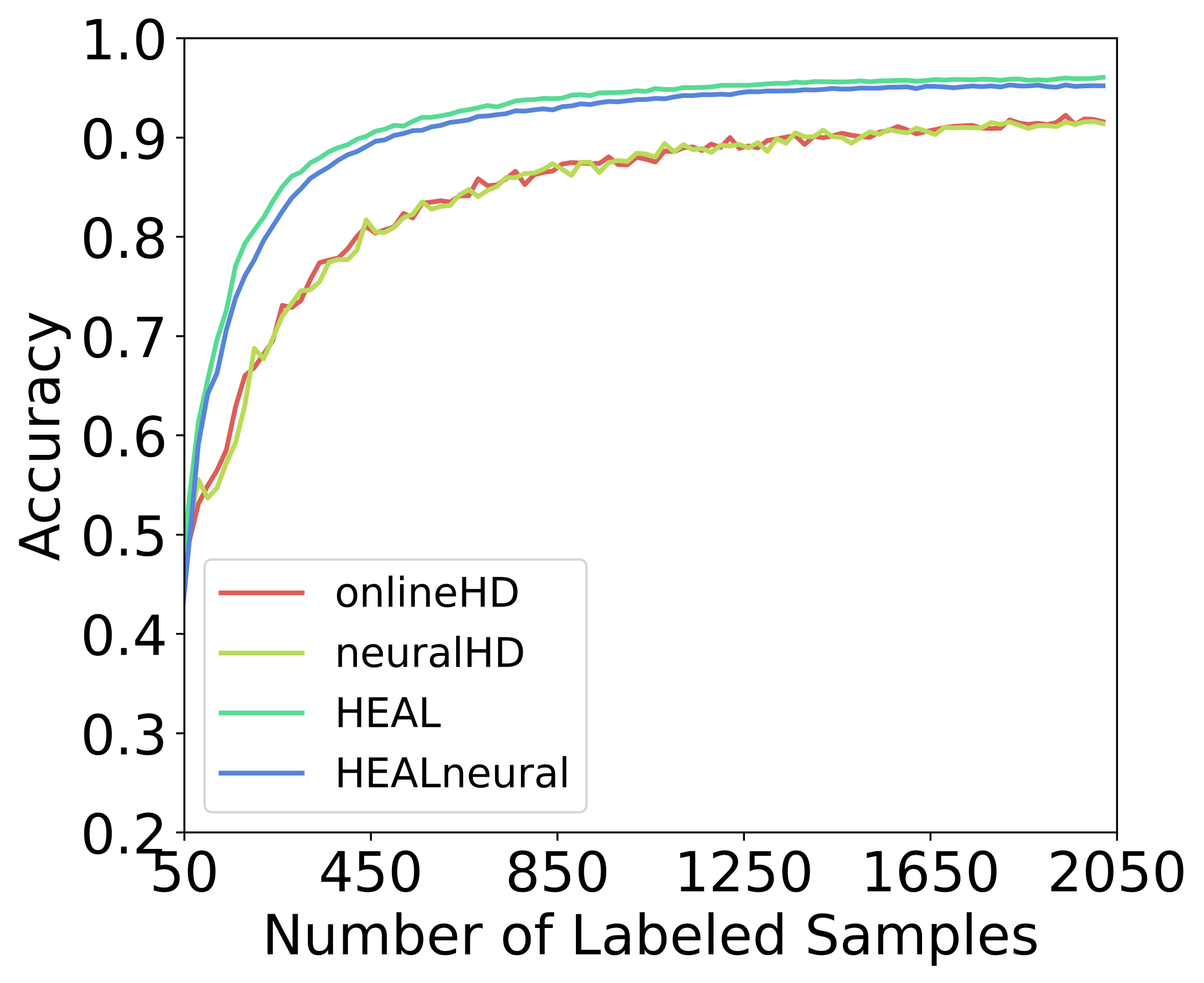}
    \caption{Learning curve comparison on ISOLET dataset between non-AL enhanced HDC classifier and the one equipped with \Design}
    \label{fig:HDC_HEAL}
\end{figure}

Prior HDC arts such as OnlineHD and NeuralHD are known for their better learning efficiency, which mainly comes from the brain-inspired hypervector representation and operations. In Fig.~\ref{fig:HDC_HEAL}, we compare the \Design and \Designn to their classifier backbones in terms of the learning curve. As for OnlineHD and NeuralHD, due to the lack of an AL framework, they will select random samples for annotation. The clear gap in the figure highlights the efficacy of the \Design and underscores the advantages of integrating the existing capabilities of HDC with a tailored AL framework to further enhance learning efficiency.

\section{Conclusion}

We introduced Hyperdimensional Efficient Active Learning (\Design), an AL framework specifically designed for HDC classification. \Design distinguishes itself by utilizing HDC-centered uncertainty and diversity-aware strategies to annotate unlabeled data efficiently. Our approach demonstrates its strength over traditional AL methods, achieving higher data efficiency and notable speedups in acquisition.

\bibliographystyle{ieeetr}
\bibliography{main}

\end{document}